\documentclass[oneside,11pt]{article}

\usepackage{times,url,mathrsfs,algorithm}
\usepackage{amsmath}
\usepackage{amsfonts}
\usepackage{graphicx}
\usepackage{subfigure}

\begin{document}

\title{Fast ABC-Boost for Multi-Class Classification}

\author{ Ping Li \\
       Department of Statistical Science\\
       Faculty of Computing and Information Science\\
       Cornell University\\
       Ithaca, NY 14853\\
       pingli@cornell.edu}
\date{}

\maketitle

\begin{abstract}
\textbf{Abc-boost} is a new line of boosting algorithms for multi-class classification, by utilizing the commonly used \textbf{sum-to-zero} constraint.  To implement {\em abc-boost},  a \textbf{base class} must be identified at each boosting step. Prior studies used a very expensive  procedure based on exhaustive search for determining the base class at each boosting step.  Good testing performance of {\em abc-boost} (implemented as \textbf{abc-mart} and \textbf{abc-logitboost}) on a variety of datasets was reported.

For large datasets, however, the exhaustive search strategy adopted in prior {\em abc-boost} algorithms can be too prohibitive. To overcome this serious limitation, this paper suggests a heuristic by introducing \textbf{Gaps} when computing the base class during training. That is, we  update the choice of the base class only for every $G$ boosting steps (i.e., $G=1$ in prior studies). We test this idea on  large datasets ({\em Covertype} and {\em Poker}) as well as datasets of moderate size. Our preliminary results are  very encouraging. On the large datasets, when $G\leq100$ (or even larger), there is essentially no loss of test accuracy compared to using $G=1$. On the moderate datasets, no obvious loss of test accuracy is observed when $G\leq 20\sim 50$. Therefore, aided by this heuristic of using {\em gaps}, it is promising that {\em abc-boost} will be a practical tool for accurate multi-class classification.

\end{abstract}

\section{Introduction}

This study focuses on significantly improving the computational efficiency of \textbf{abc-boost}, a new line of boosting algorithms recently proposed for multi-class classification~\cite{Proc:ABC_ICML09,Proc:ABC_UAI10}. Boosting~\cite{Article:Schapire_ML90,Article:Freund_95,Article:Freund_JCSS97,Article:Bartlett_AS98,Article:Schapire_ML99,Article:FHT_AS00,Proc:Mason_NIPS00,Article:Friedman_AS01,Article:Buhlmann_JASA03} has been  successful in machine learning and industry practice. 

In prior studies, {\em abc-boost} has been implemented as \textbf{abc-mart}~\cite{Proc:ABC_ICML09} and \textbf{abc-logitboost}~\cite{Proc:ABC_UAI10}. Therefore, for  completeness, we first provide a review of \textbf{logitboost}~\cite{Article:FHT_AS00} and \textbf{mart} (multiple additive regression trees)~\cite{Article:Friedman_AS01}.

\subsection{Data Probability Model and Loss Function}

We denote a training dataset by $\{y_i,\mathbf{x}_i\}_{i=1}^N$, where $N$ is the number of feature vectors (samples), $\mathbf{x}_i$ is the $i$th feature vector, and  $y_i \in \{0, 1, 2, ..., K-1\}$ is the $i$th class label, where $K\geq 3$ in multi-class classification. 

Both {\em logitboost}~\cite{Article:FHT_AS00} and {\em mart}~\cite{Article:Friedman_AS01}   can be viewed as generalizations to the classical logistic regression, which models class probabilities $p_{i,k}$ as
\begin{align}\label{eqn_logit}
p_{i,k} = \mathbf{Pr}\left(y_i = k|\mathbf{x}_i\right) = \frac{e^{F_{i,k}(\mathbf{x_i})}}{\sum_{s=0}^{K-1} e^{F_{i,s}(\mathbf{x_i})}}.
\end{align}
While logistic regression simply assumes $F_{i,k}(\mathbf{x}_i) = \beta^\text{T}_k\mathbf{x}_i$, \ {\em logitboost} and {\em mart} adopt the flexible ``additive model,''  which is a function of $M$ terms:
\begin{align}\label{eqn_F_M}
F^{(M)}(\mathbf{x}) = \sum_{m=1}^M \rho_m h(\mathbf{x};\mathbf{a}_m),
\end{align}
where  $h(\mathbf{x};\mathbf{a}_m)$, the base (weak) learner, is typically a regression tree. The parameters, $\rho_m$ and $\mathbf{a}_m$, are learned from the data, by maximizing the joint likelihood, which is equivalent to minimizing the following {\em negative log-likelihood loss} function:
\begin{align}\label{eqn_loss}
L = \sum_{i=1}^N L_i, \hspace{0.4in} L_i = - \sum_{k=0}^{K-1}r_{i,k}  \log p_{i,k}
\end{align}
where $r_{i,k} = 1$ if $y_i = k$ and $r_{i,k} = 0$ otherwise. For identifiability, $\sum_{k=0}^{K-1}F_{i,k} = 0$, i.e., the \textbf{sum-to-zero} constraint, is typically adopted
\cite{Article:FHT_AS00,Article:Friedman_AS01,Article:Zhang_JMLR04,Article:Lee_JASA04,Article:Tewari_JMLR07,Article:Zou_AOAS08,Article:Zhu_Adaboost09}.

\subsection{The (Robust) Logitboost and Mart Algorithms}

The {\em logitboost} algorithm~\cite{Article:FHT_AS00} builds the additive model (\ref{eqn_F_M}) by a greedy stage-wise  procedure, using a second-order (diagonal) approximation of the loss function (\ref{eqn_loss}). The standard practice is to implement {\em logitboost}  using regression trees. The {\em mart} algorithm~\cite{Article:Friedman_AS01} is a creative combination of gradient descent and Newton's method,  by  using the first-order information of the loss function (\ref{eqn_loss}) to construct the trees and using both the first- \& second-order derivatives  to determine the values of the terminal nodes.

Therefore, both {\em logitboost} and {\em mart} require the first two derivatives of the loss function (\ref{eqn_loss}) with respective to the function values $F_{i,k}$. \cite{Article:FHT_AS00,Article:Friedman_AS01} used the following derivatives:
\begin{align}\label{eqn_mart_d1d2}
&\frac{\partial L_i}{\partial F_{i,k}} = - \left(r_{i,k} - p_{i,k}\right),\hspace{0.3in}
\frac{\partial^2 L_i}{\partial F_{i,k}^2} = p_{i,k}\left(1-p_{i,k}\right).
\end{align}

The recent work named {\em robust logitboost}~\cite{Proc:ABC_UAI10} is a numerically stable implementation of {\em logitboost}.  \cite{Proc:ABC_UAI10} unified {\em logitboost} and {\em mart} by showing that their difference lies in the tree-split criterion for constructing the regression trees at each boosting iteration.

\subsubsection{Tree-Split Criteria for (Robust) Logitboost and Mart}\label{sec_split}

Consider $N$ weights $w_i$, and $N$ response values $z_i$, $i=1$ to $N$, which are assumed to be ordered according to the ascending order of the corresponding feature values. The tree-split procedure is to find the index $s$, $1\leq s<N$, such that the weighted square error (SE) is reduced the most if split at $s$.  That is, we seek the $s$ to maximize the \textbf{gain}: 
\begin{align}\notag
&Gain(s) = SE_{T} - (SE_{L} + SE_{R})\\\label{eqn_gain}
=&\sum_{i=1}^N (z_i - \bar{z})^2w_i - \left[
\sum_{i=1}^s (z_i - \bar{z}_L)^2w_i + \sum_{i=s+1}^N (z_i - \bar{z}_R)^2w_i\right]
\end{align}where
\begin{align}\notag
\bar{z} = \frac{\sum_{i=1}^N z_iw_i}{\sum_{i=1}^N w_i}, \hspace{0.15in}
\bar{z}_L = \frac{\sum_{i=1}^s z_iw_i}{\sum_{i=1}^s w_i},\hspace{0.15in}
\bar{z}_R = \frac{\sum_{i=s+1}^N z_iw_i}{\sum_{i=s+1}^{N} w_i}.
\end{align}
\cite{Proc:ABC_UAI10} showed the expression (\ref{eqn_gain}) can be simplified to be
\begin{align}
Gain(s) =& \frac{\left[\sum_{i=1}^s z_iw_i\right]^2}{\sum_{i=1}^s w_i}+\frac{\left[\sum_{i=s+1}^N z_iw_i\right]^2}{\sum_{i=s+1}^{N} w_i}- \frac{\left[\sum_{i=1}^N z_iw_i\right]^2}{\sum_{i=1}^N w_i}.
\end{align}

For {\em logitboost}, \cite{Article:FHT_AS00} used the weights $w_i = p_{i,k}(1-p_{i,k})$ and the responses $z_i = \frac{r_{i,k}-p_{i,k}}{p_{i,k}(1-p_{i,k})}$, i.e.,
\begin{align}\label{eqn_logit_gain}
LogitGain(s) =&  \frac{\left[\sum_{i=1}^s \left(r_{i,k} - p_{i,k}\right) \right]^2}{\sum_{i=1}^s p_{i,k}(1-p_{i,k})}+\frac{\left[\sum_{i=s+1}^N \left(r_{i,k}- p_{i,k}\right) \right]^2}{\sum_{i=s+1}^{N} p_{i,k}(1-p_{i,k})}
- \frac{\left[\sum_{i=1}^N \left(r_{i,k} - p_{i,k}\right) \right]^2}{\sum_{i=1}^N p_{i,k}(1-p_{i,k})}.
\end{align}

\vspace{0.2in}

For {\em mart}, \cite{Article:Friedman_AS01} used the weights $w_i = 1$ and the responses $z_{i,k} = r_{i,k} - p_{i,k}$, i.e.,
\begin{align}\label{eqn_mart_gain}
&MartGain(s) = \frac{1}{s} \left[\sum_{i=1}^s \left(r_{i,k} - p_{i,k}\right) \right]^2+
\frac{1}{N-s}\left[\sum_{i=s+1}^N \left(r_{i,k} - p_{i,k}\right) \right]^2-
\frac{1}{N}\left[\sum_{i=1}^N \left(r_{i,k} - p_{i,k}\right) \right]^2.
\end{align}

\subsubsection{The Robust Logitboost Algorithm}

{\scriptsize\begin{algorithm}{\small
1: $F_{i,k} = 0$, $p_{i,k} = \frac{1}{K}$, $k = 0$ to  $K-1$, $i = 1$ to $N$ \\
2: For $m=1$ to $M$ Do\\
3: \hspace{0.1in}    For $k=0$ to $K-1$ Do\\
4:  \hspace{0.15in}  $\left\{R_{j,k,m}\right\}_{j=1}^J = J$-terminal node regression tree from $\{r_{i,k} - p_{i,k}, \ \ \mathbf{x}_{i}\}_{i=1}^N$, with weights $p_{i,k}(1-p_{i,k})$ as in (\ref{eqn_logit_gain}) \\
5:   \hspace{0.2in}  $\beta_{j,k,m} = \frac{K-1}{K}\frac{ \sum_{\mathbf{x}_i \in
  R_{j,k,m}} r_{i,k} - p_{i,k}}{ \sum_{\mathbf{x}_i\in
  R_{j,k,m}}\left(1-p_{i,k}\right)p_{i,k} }$ \\
6:  \hspace{0.2in}  $F_{i,k} = F_{i,k} +
\nu\sum_{j=1}^J\beta_{j,k,m}1_{\mathbf{x}_i\in R_{j,k,m}}$ \\
7:   \hspace{0.1in} End\\
8:\hspace{0.12in} $p_{i,k} = \exp(F_{i,k})/\sum_{s=0}^{K-1}\exp(F_{i,s})$\\
9: End
\caption{\small {\em Robust logitboost}, which is very similar to the {\em mart} algorithm~\cite{Article:Friedman_AS01}, except for Line 4. }
\label{alg_robust_logitboost}}
\end{algorithm}}

Alg.~\ref{alg_robust_logitboost}  describes {\em robust logitboost} using the tree-split criterion  (\ref{eqn_logit_gain}).  In Line 6, $\nu$ is the shrinkage parameter and is normally set to be $\nu\leq 0.1$.   Note that after trees are constructed, the values of the terminal nodes are computed by
\begin{align}\label{eqn_update}
\frac{\sum_{node} z_{i,k} w_{i,k}}{\sum_{node} w_{i,k}} = \frac{\sum_{node} r_{i,k} - p_{i,k}}{\sum_{node} p_{i,k}(1-p_{i,k})},
\end{align}
which explains Line 5 of Alg. \ref{alg_robust_logitboost}.

\subsubsection{The Mart Algorithm}

The {\em mart} algorithm only uses the first derivative to construct the tree. Once the tree is constructed, \cite{Article:Friedman_AS01} applied a one-step Newton update to obtain the values of the terminal nodes. Interestingly, this one-step Newton update  yields exactly the same equation as (\ref{eqn_update}). In other words, (\ref{eqn_update}) is interpreted as weighted average in {\em logitboost} but it is interpreted as the one-step Newton update in {\em mart}. Thus, the {\em mart} algorithm is similar to Alg.~\ref{alg_robust_logitboost};  we only need to change Line 4, by replacing (\ref{eqn_logit_gain}) with (\ref{eqn_mart_gain}).

\clearpage

\section{Review Adaptive Base Class Boost (ABC-Boost)}

Developed by   \cite{Proc:ABC_ICML09}, the {\em abc-boost} algorithm consists of the following two components:
\begin{enumerate}
\item Using the widely-used sum-to-zero constraint~\cite{Article:FHT_AS00,Article:Friedman_AS01,Article:Zhang_JMLR04,Article:Lee_JASA04,Article:Tewari_JMLR07,Article:Zou_AOAS08,Article:Zhu_Adaboost09} on the loss function,  one can formulate boosting algorithms only for $K-1$ classes, by using one class as the \textbf{base class}.
\item At each boosting iteration, \textbf{adaptively} select the base class according to the training loss (\ref{eqn_loss}). \cite{Proc:ABC_ICML09} suggested an exhaustive search strategy.
\end{enumerate}

\cite{Proc:ABC_ICML09} derived the derivatives of (\ref{eqn_loss}) under the sum-to-zero constraint. Without loss of generality, we can assume that class 0 is the base class. For any $k\neq 0$,
\begin{align}\label{eqn_abc_d1}
&\frac{\partial L_i}{\partial F_{i,k}}  = \left(r_{i,0} - p_{i,0}\right) - \left(r_{i,k} - p_{i,k}\right),\\\label{eqn_abc_d2}
&\frac{\partial^2 L_i}{\partial F_{i,k}^2} = p_{i,0}(1-p_{i,0}) + p_{i,k}(1-p_{i,k}) + 2p_{i,0}p_{i,k}.
\end{align}

\cite{Proc:ABC_ICML09} combined the idea of {\em abc-boost} with {\em mart} to develop {\em abc-mart}, which achieved good performance in multi-class classification. More recently, \cite{Proc:ABC_UAI10} developed {\em abc-logitboost} by combining  {\em abc-boost} with {\em robust logitboost}.

\subsection{ABC-LogitBoost and ABC-Mart}

Alg.~\ref{alg_abc-logitboost} presents  {\em abc-logitboost}, using the derivatives in (\ref{eqn_abc_d1}) and (\ref{eqn_abc_d2})  and the same exhaustive search strategy proposed in~\cite{Proc:ABC_ICML09}. Compared to  Alg.~\ref{alg_robust_logitboost}, {\em abc-logitboost} differs from {\em (robust) logitboost} in that they use different derivatives and {\em abc-logitboost} needs an additional loop to select the base class at each boosting iteration.

\begin{algorithm}[h]{\small
1: $F_{i,k} = 0$,\ \  $p_{i,k} = \frac{1}{K}$, \ \ \ $k = 0$ to  $K-1$, \ $i = 1$ to $N$ \\
2: For $m=1$ to $M$ Do\\
3: \hspace{0.1in}    For $b=0$ to $K-1$, Do\\
4: \hspace{0.2in}    For $k=0$ to $K-1$, $k\neq b$, Do\\
5:  \hspace{0.3in}  $\left\{R_{j,k,m}\right\}_{j=1}^J = J$-terminal
node regression tree from
  $\{-(r_{i,b} - p_{i,b}) +  (r_{i,k} - p_{i,k}), \ \ \mathbf{x}_{i}\}_{i=1}^N$ with\\
:\hspace{0.4in} weights $p_{i,b}(1-p_{i,b})+p_{i,k}(1-p_{i,k})+2p_{i,b}p_{i,k}$, in Sec.~\ref{sec_split}.
 \\
6:   \hspace{0.3in}  $\beta_{j,k,m} = \frac{ \sum_{\mathbf{x}_i \in
  R_{j,k,m}} -(r_{i,b} - p_{i,b}) + (r_{i,k} - p_{i,k})  }{ \sum_{\mathbf{x}_i\in
  R_{j,k,m}} p_{i,b}(1-p_{i,b})+ p_{i,k}\left(1-p_{i,k}\right) + 2p_{i,b}p_{i,k} }$ \\\\
7:  \hspace{0.3in}  $g_{i,k,b} = F_{i,k} +
\nu\sum_{j=1}^J\beta_{j,k,m}1_{\mathbf{x}_i\in R_{j,k,m}}$ \\
8:   \hspace{0.2in} End\\
9: \hspace{0.2in} $g_{i,b,b} = - \sum_{k\neq b} g_{i,k,b}$ \\
10: \hspace{0.2in}  $q_{i,k} = \exp(g_{i,k,b})/\sum_{s=0}^{K-1}\exp(g_{i,s,b})$ \\
11: \hspace{0.2in} $L^{(b)} = -\sum_{i=1}^N \sum_{k=0}^{K-1} r_{i,k}\log\left(q_{i,k}\right)$\\
12: \hspace{0.1in} End\\
13: \hspace{0.1in} $B(m) = \underset{b}{\text{argmin}} \  \ L^{(b)}$\\
14: \hspace{0.1in} $F_{i,k} = g_{i,k,B(m)}$\\
15:\hspace{0.1in}  $p_{i,k} = \exp(F_{i,k})/\sum_{s=0}^{K-1}\exp(F_{i,s})$ \\
16: End}
\caption{{\em Abc-logitboost} using the exhaustive search strategy for the base class, as suggested in \cite{Proc:ABC_ICML09}.  The vector $B$ stores the base class numbers. }
\label{alg_abc-logitboost}
\end{algorithm}%

Again, {\em abc-logitboost} differs from {\em abc-mart} only in the tree-split procedure (Line 5 in Alg.~\ref{alg_abc-logitboost}).

\subsection{Why Does the Choice of Base Class Matter?}

\cite{Proc:ABC_UAI10} used the Hessian matrix, to demonstrate why the choice of the base class matters.

The chose of the base class matters because of the diagonal approximation; that is, fitting a regression tree for each class at each boosting iteration. To see this, we can take a look at the Hessian matrix, for $K=3$. Using the original logitboost/mart derivatives (\ref{eqn_mart_d1d2}), the determinant of the Hessian matrix is  \begin{align}\notag
&\left|\begin{array}{ccc}
\frac{\partial^2 L_i}{\partial p_{0}^2} & \frac{\partial^2 L_i}{\partial p_{0} p_{1}} & \frac{\partial^2 L_i}{\partial p_{0} p_{2}}\\
\frac{\partial^2 L_i}{\partial p_{1}p_{0}} & \frac{\partial^2 L_i}{\partial p_{1}^2} & \frac{\partial^2 L_i}{\partial p_{1} p_{2}}\\
\frac{\partial^2 L_i}{\partial p_{2}p_{0}} & \frac{\partial^2 L_i}{\partial p_{2}p_{1}} & \frac{\partial^2 L_i}{\partial p_{2}^2}
\end{array}\right|
 =
\left|\begin{array}{ccc}
p_0(1-p_0) & -p_0p_1& -p_0p_2\\
-p_1p_0 &p_1(1-p_1) & -p_1p_2\\
-p_2p_0 &-p_2p_1 &p_2(1-p_2)\\
\end{array}\right|
={ 0}
\end{align}
as expected, because there are  only $K-1$ degrees of freedom. A simple fix is to use the diagonal approximation~\cite{Article:FHT_AS00,Article:Friedman_AS01}. In fact, when  trees are used as the weak learner, it seems one must use the diagonal approximation.

Now, consider the derivatives (\ref{eqn_abc_d1}) and (\ref{eqn_abc_d2}) used in {\em abc-mart} and {\em abc-logitboost}. This time, when $K=3$ and $k=0$ is the base class,  we only have a 2 by 2 Hessian matrix, whose determinant is
\begin{align}\notag
&\left|\begin{array}{cc}
\frac{\partial^2 L_i}{\partial p_{1}^2} & \frac{\partial^2 L_i}{\partial p_{1} p_{2}}\\
\frac{\partial^2 L_i}{\partial p_{2}p_{1}} & \frac{\partial^2 L_i}{\partial p_{2}^2}
\end{array}\right|=
\left|\begin{array}{cc}
p_0(1-p_0)+p_1(1-p_1)+2p_0p_1 & p_0-p_0^2+p_0p_1+p_0p_2-p_1p_2\\
p_0-p_0^2+p_0p_1+p_0p_2-p_1p_2 &p_0(1-p_0)+p_2(1-p_2)+2p_0p_2  \\
\end{array}\right|\\\notag
=&p_0p_1 + p_0p_2 + p_1p_2 - p_0p_1^2-p_0p_2^2-p_1p_2^2-p_2p_1^2-p_1p_0^2-p_2p_0^2
+6p_0p_1p_2,
\end{align}

which is non-zero and is in fact independent of the choice of the base class (even though we assume $k=0$ as the base in this example).  In other words, the choice of the base class would not matter if the full Hessian is used.

However, because we will have to use diagonal approximation in order to construct trees at each iteration, the choice of the base class will matter.

\subsection{Datasets Used for Testing Fast ABC-Boost}

We will test \textbf{fast abc-boost} using a subset of the datasets in~\cite{Proc:ABC_UAI10}, as listed in Table \ref{tab_data}. Because the computational cost of {\em abc-boost} is not a concern for small datasets, this study focuses on fairly large datasets ({\em Covertype} and {\em Poker}) as well as datasets of moderate size ({\em Mnist10k} and {\em M-Image}).

\begin{table}[h]
\caption{Datasets}
\begin{center}{\small
\begin{tabular}{l r r r r}
\hline \hline
dataset &$K$ & \# training & \# test &\# features\\
\hline
Covertype290k &7 & 290506 & 290506 & 54\\
Poker525k &10 & 525010 &500000 &25\\
Poker275k &10 & 275010 &500000 &25\\
Mnist10k &10 &10000 &60000&784\\
M-Image &10 &12000 &50000&784\\
\hline\hline
\end{tabular}
}
\end{center}
\label{tab_data}\vspace{-0.15in}
\end{table}

\subsection{Review the Detailed Experiment Results of ABC-Boost on Mnist10k and M-Image}

For these  two datasets, \cite{Proc:ABC_UAI10} experimented with every combination of $J\in\{4, 6, 8, 10, 12, 14, 16,18, 20, 24, 30, 40, 50\}$ and $\nu \in\{0.04, 0.06, 0.08, 0.1\}$. The four boosting algorithms were trained till the training loss (\ref{eqn_loss}) was close to the machine accuracy, to exhaust the capacity of the learners, for reliable comparisons, up to $M=10000$ iterations. Since no obvious overfitting was observed, the test mis-classification errors at the last iterations were reported.

Table~\ref{tab_Mnist10k} and Table~\ref{tab_M-Image}  present the test mis-classification errors, which verify the consistent improvements of (A) {\em abc-logitboost} over {\em (robust) logitboost}, (B) {\em abc-logitboost} over {\em abc-mart}, (C) {\em (robust) logitboost} over {\em mart}, and (D) {\em abc-mart} over {\em mart}. The tables also verify that the performances are not too sensitive to the parameters ($J$ and $\nu$).

\begin{table}[h]
\caption{\small\textbf{\em Mnist10k}. Upper table: The test mis-classification errors of  {\em mart} and \textbf{\em abc-mart} (bold numbers). Bottom table: The test  errors of  {\em logitboost} and \textbf{\em abc-logitboost} (bold numbers)}
\begin{center}
{\small
{\begin{tabular}{l l l l l }
\hline \hline

&{\em mart} & \textbf{\em abc-mart}\\\hline
  &$\nu = 0.04$ &$\nu=0.06$ &$\nu=0.08$ &$\nu=0.1$ \\
\hline

$J=4$        &3356        \textbf{3060}        &3329        \textbf{3019}        &3318        \textbf{2855}        &3326        \textbf{2794}\\
$J=6$         &3185        \textbf{2760}        &3093        \textbf{2626}        &3129        \textbf{2656}        &3217        \textbf{2590}\\
$J=8$         &3049        \textbf{2558}        &3054        \textbf{2555}        &3054        \textbf{2534}        &3035        \textbf{2577}\\
$J=10$         &3020        \textbf{2547}        &2973        \textbf{2521}        &2990        \textbf{2520}        &2978        \textbf{2506}\\
$J=12$         &2927        \textbf{2498}        &2917        \textbf{2457}        &2945        \textbf{2488}       &2907        \textbf{2490}\\
$J=14$         &2925        \textbf{2487}        &2901        \textbf{2471}        &2877        \textbf{2470}       &2884        \textbf{2454}\\
$J=16$         &2899        \textbf{2478}        &2893        \textbf{2452}        &2873        \textbf{2465}       &2860        \textbf{2451}\\
$J=18$         &2857        \textbf{2469}        &2880        \textbf{2460}        &2870        \textbf{2437}        &2855        \textbf{2454}\\
$J=20$         &2833        \textbf{2441}        &2834        \textbf{2448}        &2834        \textbf{2444} &2815        \textbf{2440}\\
$J=24$         &2840        \textbf{2447}        &2827        \textbf{2431}        &2801        \textbf{2427}        &2784        \textbf{2455}\\
$J=30$         &2826        \textbf{2457}        &2822        \textbf{2443}        &2828        \textbf{2470}        &2807        \textbf{2450}\\
$J=40$         &2837        \textbf{2482}        &2809        \textbf{2440}        &2836        \textbf{2447}        &2782        \textbf{2506}\\
$J=50$         &2813        \textbf{2502}        &2826        \textbf{2459}        &2824        \textbf{2469}        &2786        \textbf{2499}\\
\hline \hline
 &{\em logitboost} & \textbf{\em abc-logit}\\\hline
  &$\nu = 0.04$ &$\nu=0.06$ &$\nu=0.08$ &$\nu=0.1$ \\
\hline

$J=4$        &2936        \textbf{2630}        &2970        \textbf{2600}        &2980        \textbf{2535}        &3017        \textbf{2522}\\
$J=6$        &2710        \textbf{2263}        &2693        \textbf{2252}        &2710        \textbf{2226}       &2711        \textbf{2223}\\
$J=8$        &2599        \textbf{2159}        &2619        \textbf{2138}        &2589        \textbf{2120}       &2597        \textbf{2143}\\
$J=10$        &2553        \textbf{2122}        &2527        \textbf{2118}       &2516        \textbf{2091}       &2500        \textbf{2097}\\
$J=12$        &2472        \textbf{2084}        &2468        \textbf{2090}       &2468        \textbf{2090}       &2464        \textbf{2095}\\
$J=14$        &2451        \textbf{2083}        &2420        \textbf{2094}       &2432        \textbf{2063}       &2419        \textbf{2050}\\
$J=16$        &2424        \textbf{2111}        &2437        \textbf{2114}       &2393        \textbf{2097}       &2395        \textbf{2082}\\
$J=18$        &2399        \textbf{2088}        &2402        \textbf{2087}       &2389        \textbf{2088}       &2380        \textbf{2097}\\
$J=20$        &2388        \textbf{2128}        &2414        \textbf{2112}       &2411        \textbf{2095}       &2381        \textbf{2102}\\
$J=24$        &2442        \textbf{2174}        &2415        \textbf{2147}       &2417       \textbf{2129}      &2419        \textbf{2138}\\
$J=30$       & 2468        \textbf{2235}        &2434        \textbf{2237}       &2423        \textbf{2221}        &2449        \textbf{2177}\\
$J=40$       & 2551        \textbf{2310}        &2509        \textbf{2284}       &2518        \textbf{2257}        &2531        \textbf{2260}\\
$J=50$       & 2612        \textbf{2353}        &2622        \textbf{2359}       &2579        \textbf{2332}        &2570        \textbf{2341}\\
\hline\hline
\end{tabular}}}
\end{center}
\label{tab_Mnist10k}
\end{table}

\begin{table}[h]
\caption{\small\textbf{\em M-Image}. Upper table: The test mis-classification errors of  {\em mart} and \textbf{\em abc-mart} (bold numbers). Bottom table: The test of  {\em logitboost} and \textbf{\em abc-logitboost} (bold numbers)}
\begin{center}
{\small
{\begin{tabular}{l l l l l }
\hline \hline

&{\em mart} & \textbf{\em abc-mart}\\\hline
  &$\nu = 0.04$ &$\nu=0.06$ &$\nu=0.08$ &$\nu=0.1$ \\
\hline

$J=4$  &6536   \textbf{5867}  &6511  \textbf{5813}  &6496  \textbf{5774}   &6449 \textbf{5756}\\
$J=6$  &6203   \textbf{5471}  &6174  \textbf{5414}  &6176  \textbf{5394}   &6139 \textbf{5370}\\
$J=8$  &6095   \textbf{5320}  &6081  \textbf{5251}  &6132  \textbf{5141}   &6220 \textbf{5181}\\
$J=10$  &6076   \textbf{5138}  &6104  \textbf{5100}  &6154  \textbf{5086}   &6332 \textbf{4983}\\
$J=12$  &6036   \textbf{4963}  &6086  \textbf{4956}  &6104  \textbf{4926}   &6117 \textbf{4867}\\
$J=14$  &5922   \textbf{4885}  &6037  \textbf{4866}  &6018  \textbf{4789}  &5993 \textbf{4839}\\
$J=16$  &5914   \textbf{4847}  &5937  \textbf{4806}  &5940  \textbf{4797}   &5883 \textbf{4766}\\
$J=18$  &5955   \textbf{4835}  &5886  \textbf{4778}  &5896  \textbf{4733}   &5814 \textbf{4730}\\
$J=20$  &5870   \textbf{4749}  &5847  \textbf{4722}  &5829  \textbf{4707}   &5821 \textbf{4727}\\
$J=24$  &5816   \textbf{4725}  &5766  \textbf{4659}  &5785  \textbf{4662}   &5752 \textbf{4625}\\
$J=30$  &5729   \textbf{4649}  &5738  \textbf{4629}  &5724  \textbf{4626}   &5702 \textbf{4654}\\
$J=40$  &5752   \textbf{4619}  &5699  \textbf{4636}  &5672  \textbf{4597}   &5676 \textbf{4660}\\
$J=50$  &5760   \textbf{4674}  &5731  \textbf{4667}  &5723  \textbf{4659}   &5725 \textbf{4649}\\
\hline \hline
 &{\em logitboost} & \textbf{\em abc-logit}\\\hline
  &$\nu = 0.04$ &$\nu=0.06$ &$\nu=0.08$ &$\nu=0.1$ \\
\hline

$J=4$  &5837   \textbf{5539}  &5852  \textbf{5480}  &5834  \textbf{5408}   &5802 \textbf{5430}\\
$J=6$  &5473   \textbf{5076}  &5471  \textbf{4925}  &5457  \textbf{4950}   &5437 \textbf{4919}\\
$J=8$  &5294   \textbf{4756}  &5285  \textbf{4748}  &5193  \textbf{4678}   &5187 \textbf{4670}\\
$J=10$  &5141   \textbf{4597} &5120  \textbf{4572}  &5052  \textbf{4524}   &5049 \textbf{4537}\\
$J=12$  &5013  \textbf{4432}  &5016  \textbf{4455}  &4987  \textbf{4416}   &4961 \textbf{4389}\\
$J=14$  &4914   \textbf{4378}  &4922  \textbf{4338}  &4906  \textbf{4356}  &4895 \textbf{4299}\\
$J=16$  &4863   \textbf{4317}  &4842  \textbf{4307}  &4816  \textbf{4279}   &4806 \textbf{4314}\\
$J=18$  &4762   \textbf{4301}  &4740  \textbf{4255}  &4754  \textbf{4230}   &4751 \textbf{4287}\\
$J=20$  &4714   \textbf{4251}  &4734  \textbf{4231}  &4693  \textbf{4214}   &4703 \textbf{4268}\\
$J=24$  &4676   \textbf{4242}  &4610  \textbf{4298}  &4663  \textbf{4226}   &4638 \textbf{4250}\\
$J=30$  &4653   \textbf{4351}  &4662  \textbf{4307}  &4633  \textbf{4311}   &4643 \textbf{4286}\\
$J=40$  &4713   \textbf{4434}  &4724  \textbf{4426}  &4760  \textbf{4439}   &4768 \textbf{4388}\\
$J=50$  &4763   \textbf{4502}  &4795  \textbf{4534}  &4792  \textbf{4487}   &4799 \textbf{4479}\\
\hline\hline
\end{tabular}}}
\end{center}
\label{tab_M-Image}
\end{table}

\clearpage

\section{Fast ABC-Boost}

Recall that, in {\em abc-boost}, the base class must be identified at each boosting iteration. The exhaustive search strategy used in~\cite{Proc:ABC_ICML09, Proc:ABC_UAI10} is obviously very expensive. In this paper, our main contribution  is a proposal for speeding up {\em abc-boost} by introducing \textbf{Gaps} when selecting the base class. Again, we illustrate our strategy using {\em abc-mart} and {\em abc-logitboost}, which are only two implementations of {\em abc-boost} so far. \\

Assuming $M$ boosting iterations, the computation cost of {\em mart} and {\em logitboost} is $O(KM)$. However,  the computation cost of {\em abc-mart} and {\em abc-logitboost} $O\left(K(K-1)M\right)$, which can be prohibitive.\\

The reason we need to select the {\em base class} is because we have to use the the diagonal approximation in order to fit a regression separately for each class at every boosting iteration.  Based on this insight, we really do not have to re-compute the base class for every iteration. Instead, we only compute the base class for every $G$ steps, where $G$ is the {\em gap}  and $G=1$ means we select the base class for every iteration. \\

After introducing {\em gaps}, the computation cost of {\em fast abc-boost} is reduced to $O\left(K(K-1)\frac{M}{G} + \left(M-\frac{M}{G}\right)(K-1)\right)$. One can verify that when $G = (K-1)$, the cost of {\em fast abc-boost} is at most twice as the cost of {\em logitboost}. As we increases $G$ more, the additional computational overhead of {\em fast abc-boost} further diminishes. \\

The parameter $G$ can be viewed as a new tuning parameter. Our experiments (in the following subsections) illustrate  that when $G\leq 100$ (or $G\leq 20\sim 50$), there would be no obvious loss of test accuracies in large datasets (or moderate datasets).

\vspace{-0.1in}
\subsection{Experiments on Large Datasets, {\em Poker525k}, {\em Poker275k}, and {\em Covertype290k}}

\begin{figure}[h]
\begin{center}\mbox{
\includegraphics[width = 2.2in]{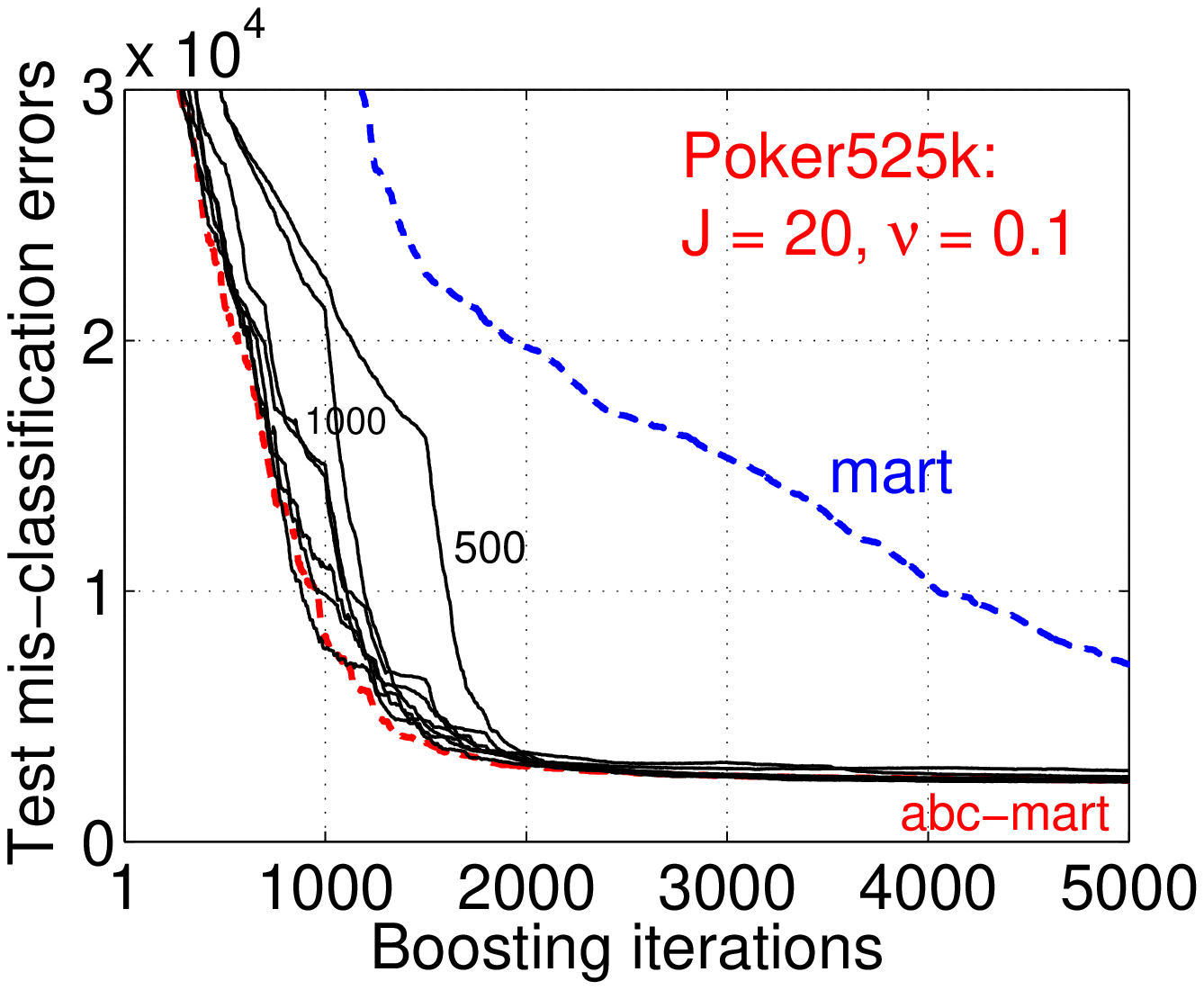}\hspace{-0.1in}
\includegraphics[width = 2.2in]{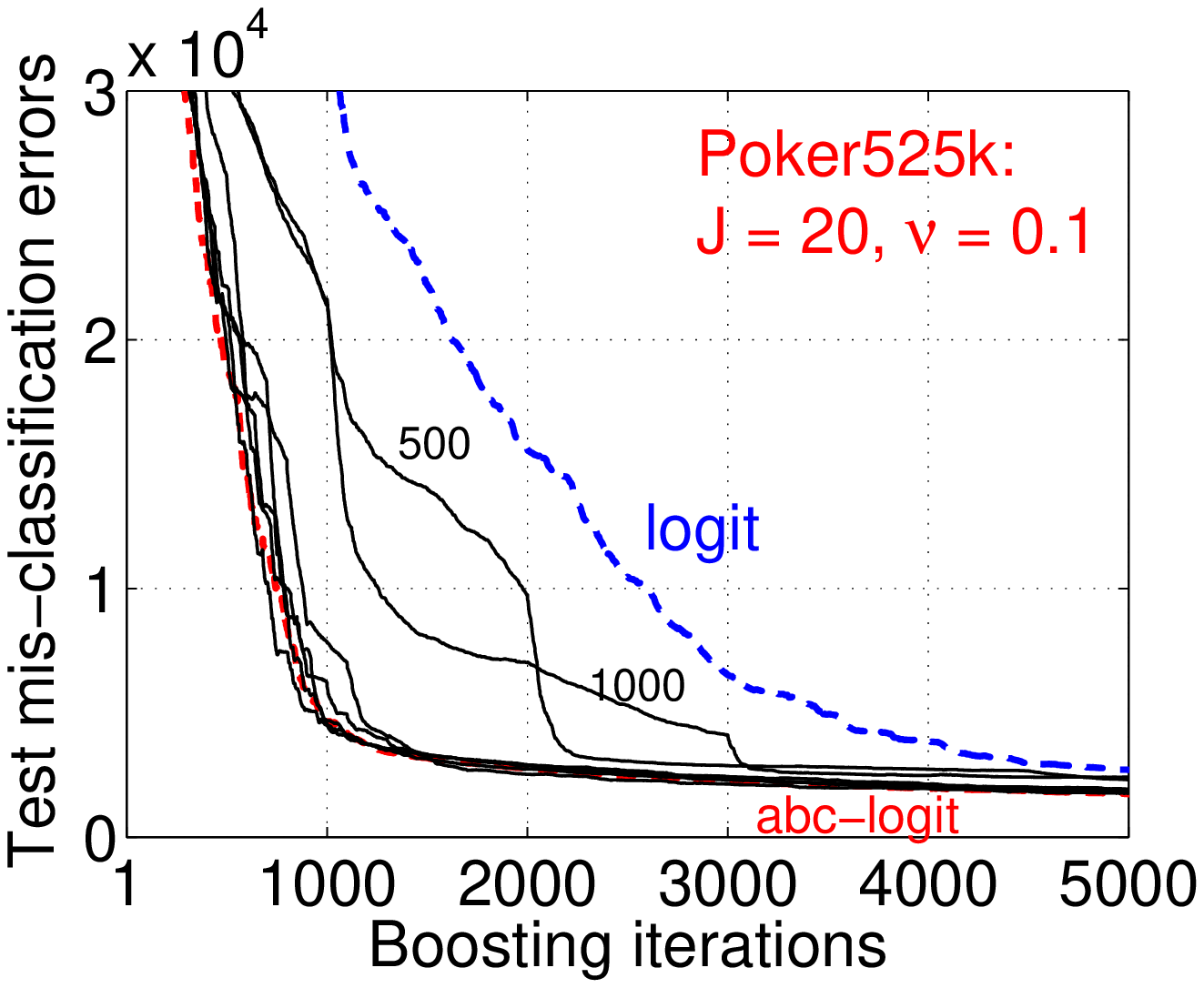}\hspace{-0.1in}
\includegraphics[width = 2.2in]{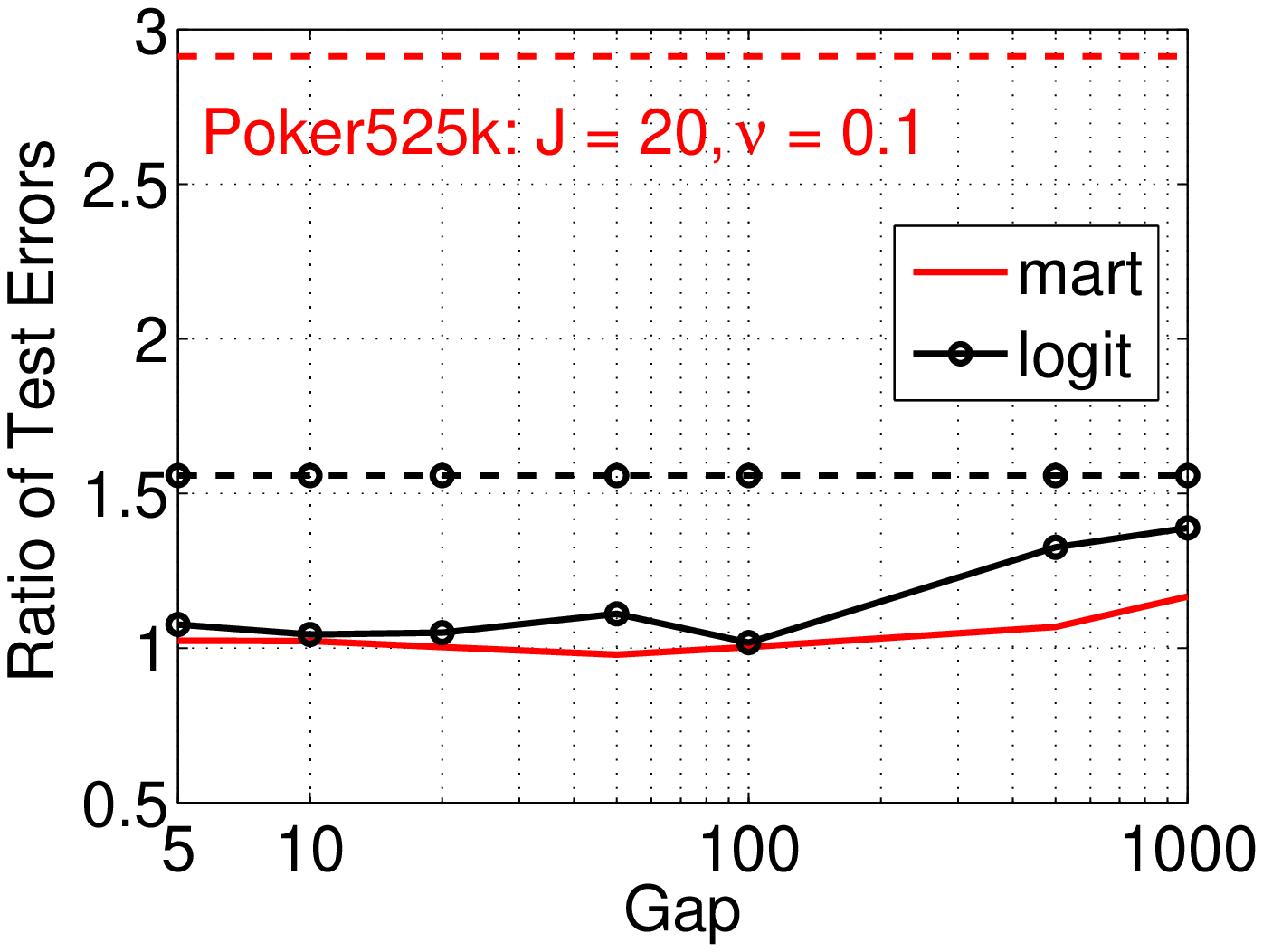}

}
\end{center}\vspace{-0.2in}
\caption{\textbf{Poker525k}\hspace{0.2in}  \textbf{Left panel}: test mis-classification errors of {\em abc-mart} (with $G=1, 5, 10, 20, 50, 100, 500, 1000$) and {\em mart}, for all boosting iterations up to $M=5000$ steps. We only label the curves which are distinguishable (in this case $G=500$ and 1000). \textbf{Middle panel}: test mis-classification errors of {\em abc-logitboost} and {\em (robust) logitboost}. Note that, at $M=5000$, the test error of {\em abc-logitboost} is significantly smaller than the test error of {\em logitboost}, even though, due to the scaling issue, the difference may be less obvious in the figure. \textbf{Right panel}: the ratios of test errors, i.e., {\em mart} over {\em abc-mart} and {\em logitboost} over {\em abc-logitboost}, at the last (i.e., $M=5000$) boosting iteration. The two \textbf{dashed horizontal lines} represent the test error ratios at $G=1$ (i.e., the original {\em abc-boost}). Note that a ratio of 1.5 (or even 3) should be considered extremely large for classification tasks.} \label{fig_Poker525k}
\end{figure}

As presented in~\cite{Proc:ABC_UAI10}, on the {\em Poker} dataset, {\em abc-boost} achieved very remarkable improvements over {\em mart} and {\em logitboost}, especially when the number of boosting iterations was not too large. In fact, even at $M=5000$ iterations, the mis-classification error of {\em mart} (or {\em (robust) logitboost}) is 3 times (or 1.5 times) as large as the error of {\em abc-mart} (or {\em abc-logitboost}); see the rightmost panel of Figure~\ref{fig_Poker525k}.

For all datasets, we experiment with $G=1$ (i.e., the original {\em abc-boost}), $5, 10, 20, 50, 100, 500, 1000$.  As shown in Figure~\ref{fig_Poker525k}, using {\em fast abc-boost} with $G\leq 100$, there is no obvious loss of test accuracies on {\em Poker525k}. In fact, using {\em abc-mart}, even with $G=1000$, there is only very little loss of accuracy.\\

Note that it is possible for {\em fast abc-boost} to achieve smaller test errors than {\em abc-boost}; for example, the ratios of test errors in the right panel of Figure~\ref{fig_Poker525k} may be below 1.0. This interesting phenomenon is not surprising. After all, $G$ can be viewed as  tuning parameter and using $G>1$ may have some {\em regularization} effect because that would be less greedy. \\

\begin{figure}[h]
\begin{center}\mbox{
\includegraphics[width = 2.2in]{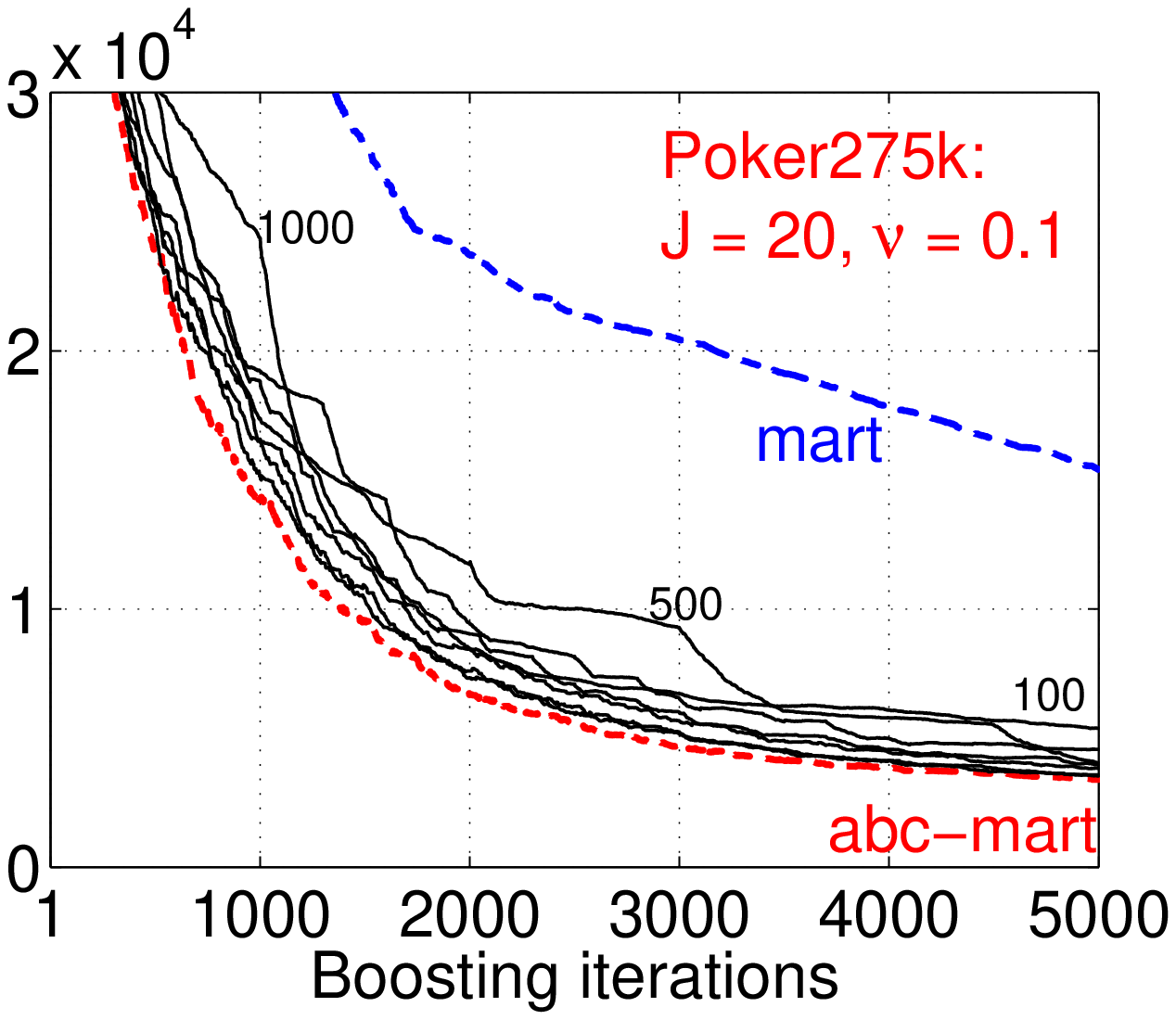}\hspace{-0.1in}
\includegraphics[width = 2.2in]{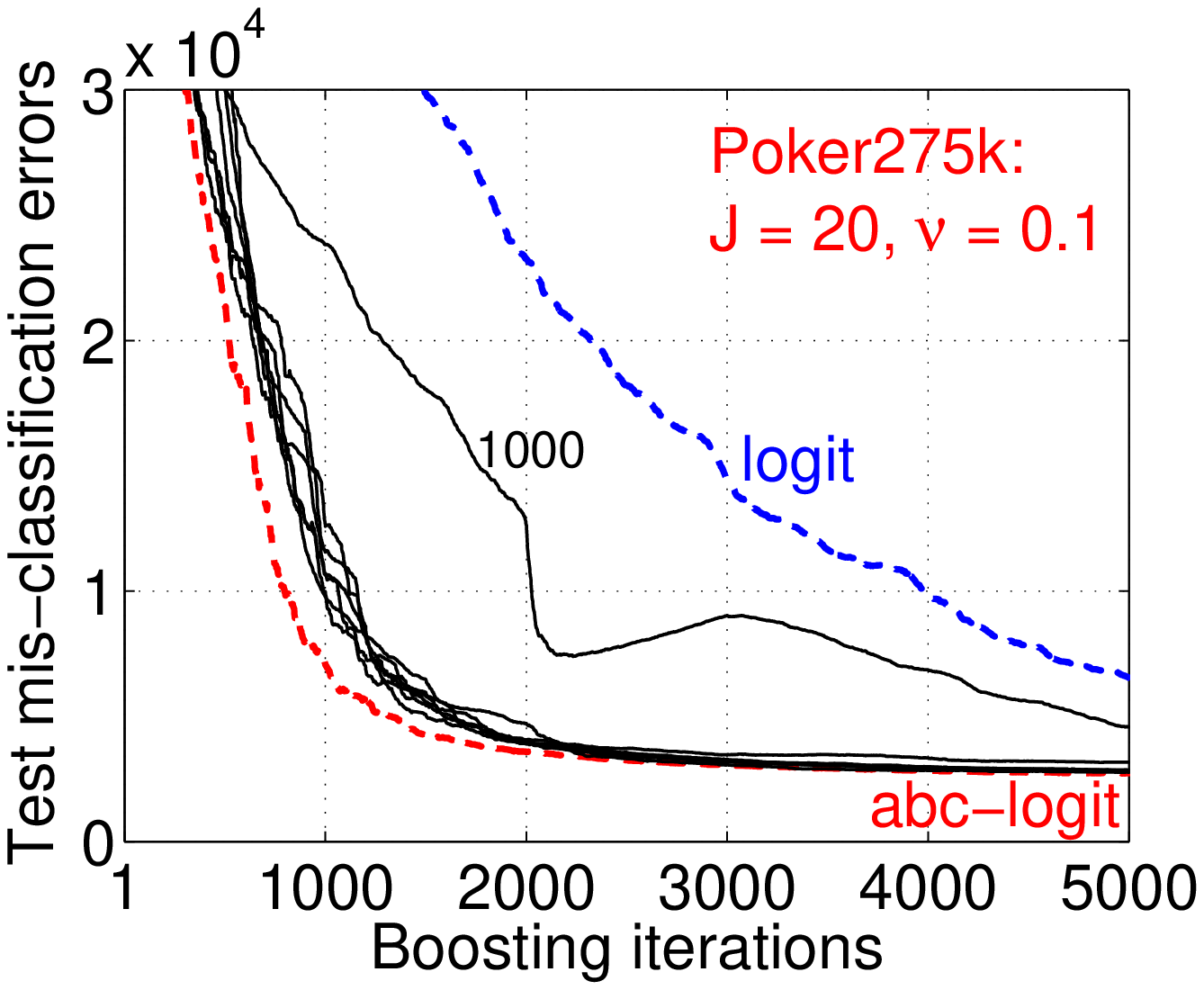}\hspace{-0.1in}
\includegraphics[width = 2.2in]{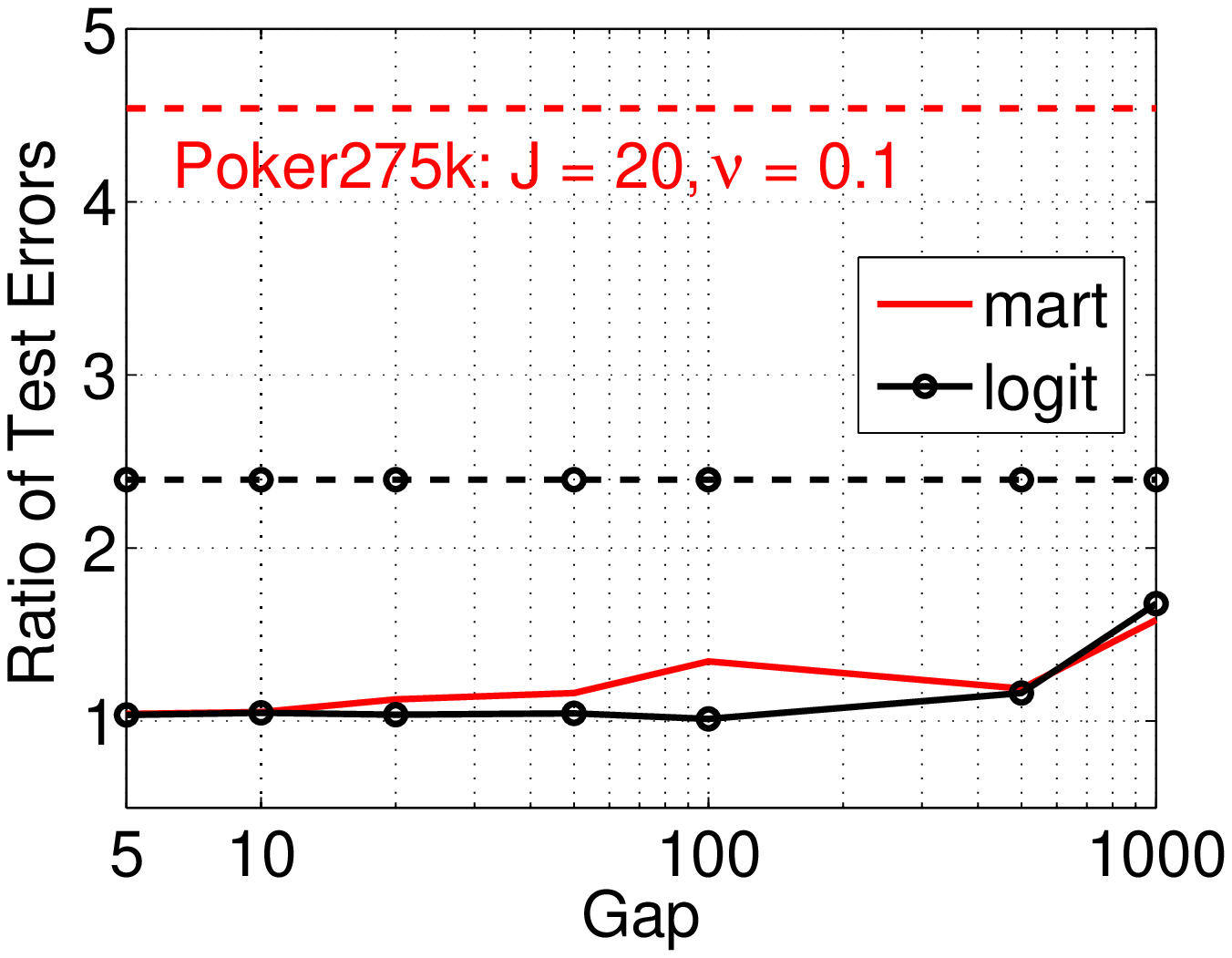}

}
\end{center}\vspace{-0.3in}
\caption{\textbf{Poker275k}. \hspace{0.2in} See the caption of Figure~\ref{fig_Poker525k} for explanations.}\label{fig_Poker275k}
\end{figure}

Figure~\ref{fig_Poker275k} presents the test error results on {\em Poker275k}, which are very similar to the results on {\em Poker525k}.

\vspace{0.1in}

\begin{figure}[h]
\begin{center}\mbox{
\includegraphics[width = 2.2in]{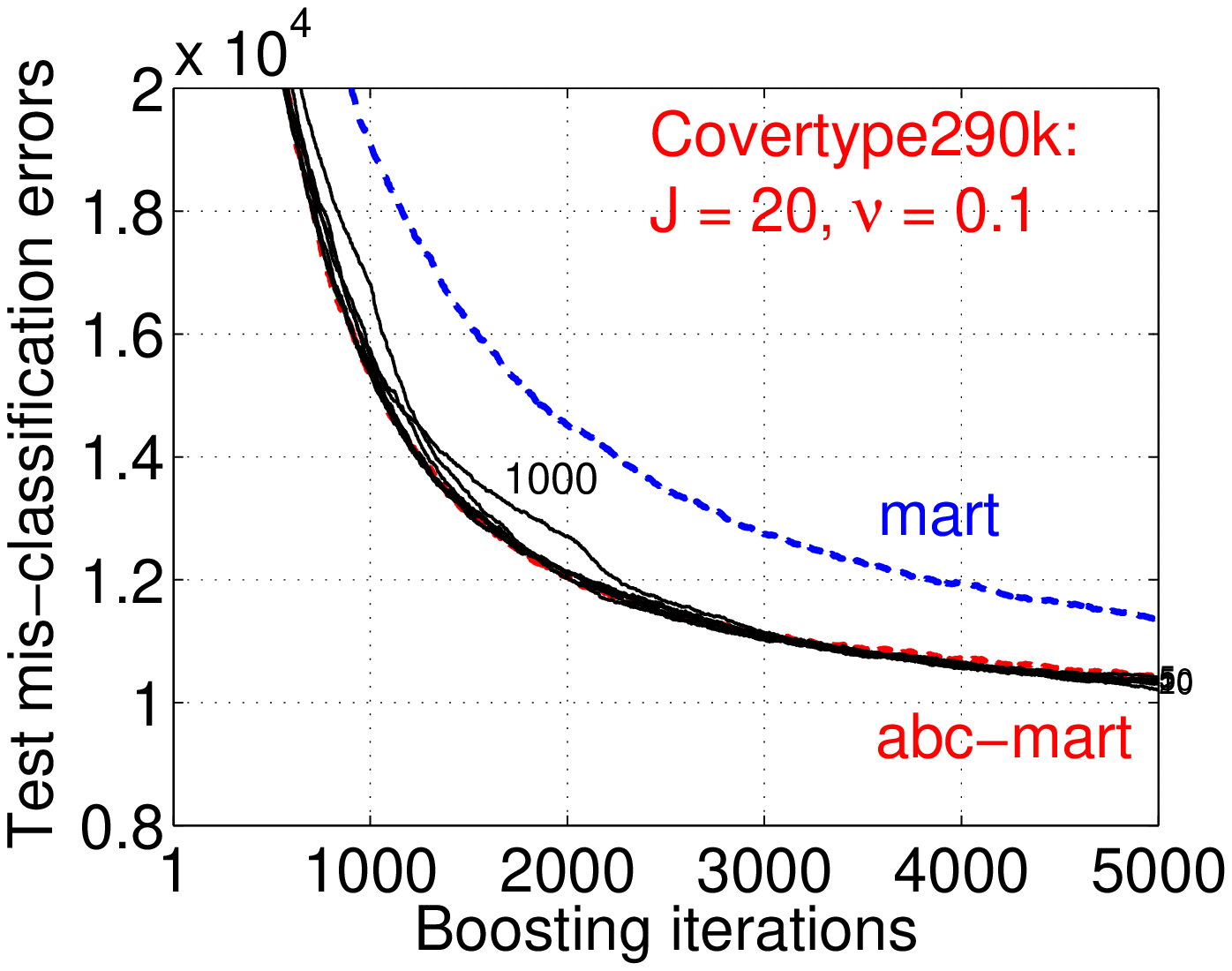}\hspace{-0.1in}
\includegraphics[width = 2.2in]{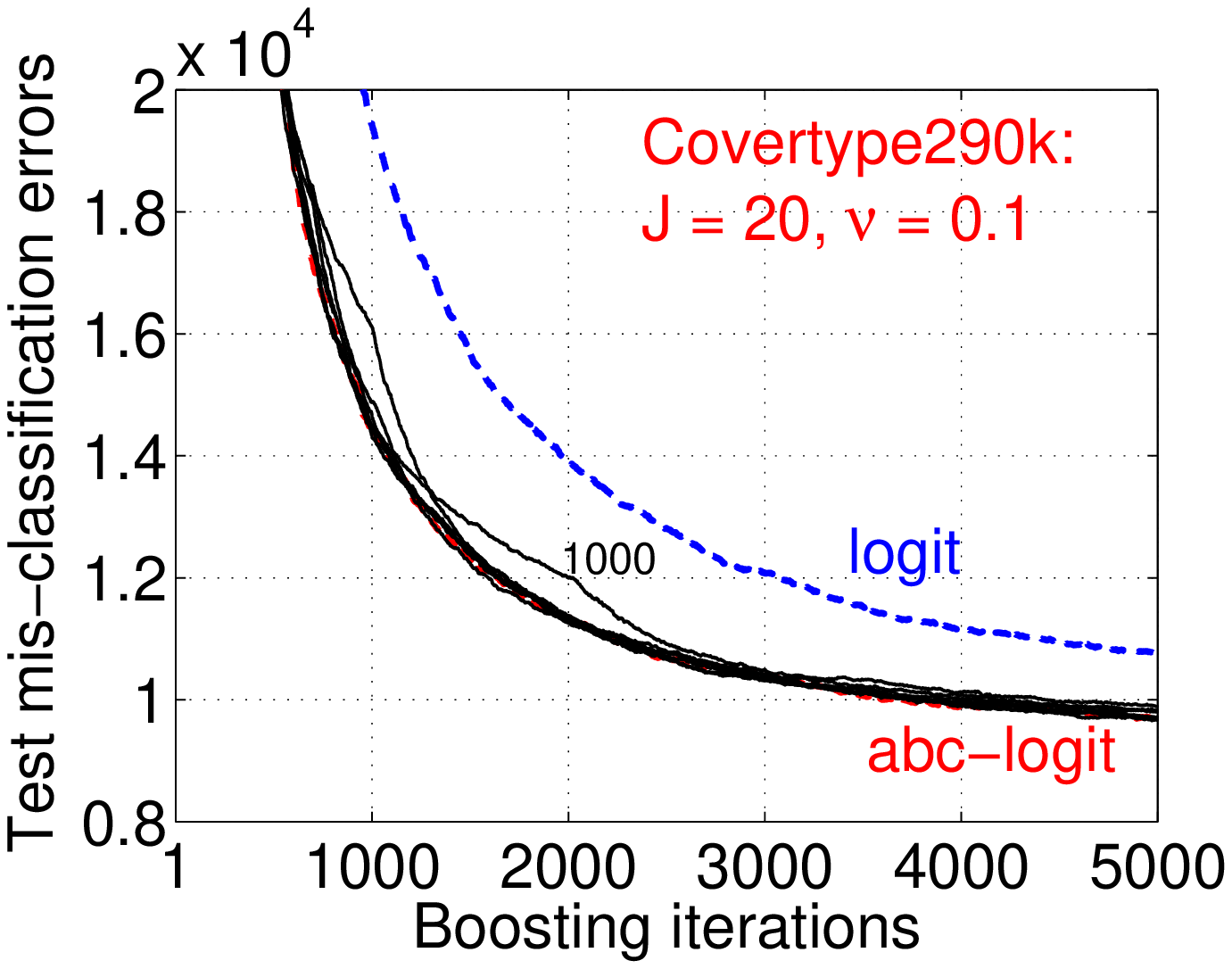}\hspace{-0.1in}
\includegraphics[width = 2.2in]{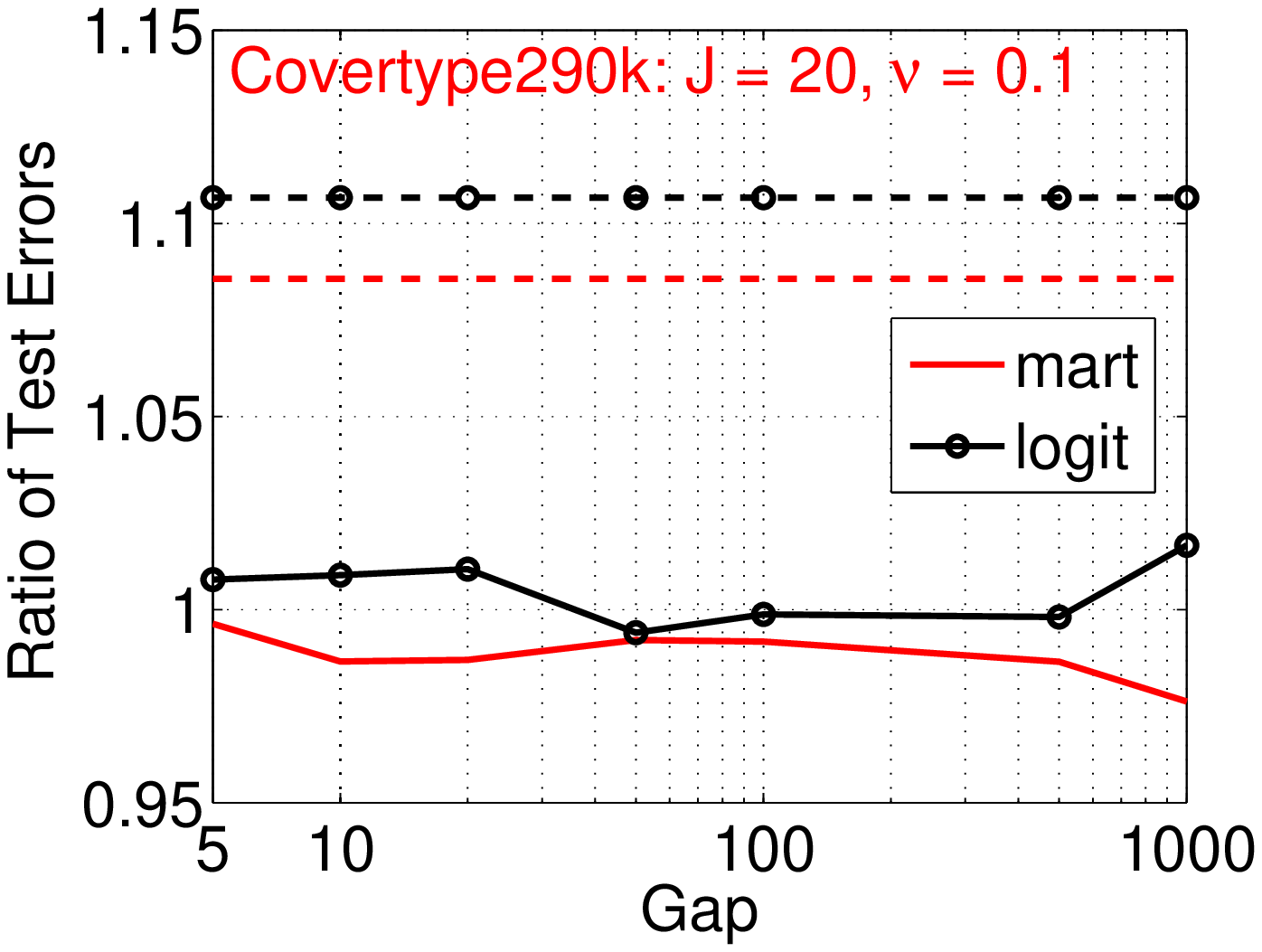}

}
\end{center}\vspace{-0.3in}
\caption{\textbf{Covertype290k}}\label{fig_Covertype290k}
\end{figure}

Figure~\ref{fig_Covertype290k} presents the test error results on {\em Covertype290k}. For this dataset, even with $G=1000$, we notice essentially no loss of test accuracies.

\subsection{Experiments on Moderate Datasets, {\em M-Image} and {\em Mnist10k}}

The situation is somewhat different on datasets that are not too large.  Recall, for these two datasets, we terminate the training if the training loss (\ref{eqn_loss}) is to close to the machine accuracy, up to $M=10000$ iterations.

\begin{figure}[h!]
\begin{center}
\mbox{
\includegraphics[width = 2.2in]{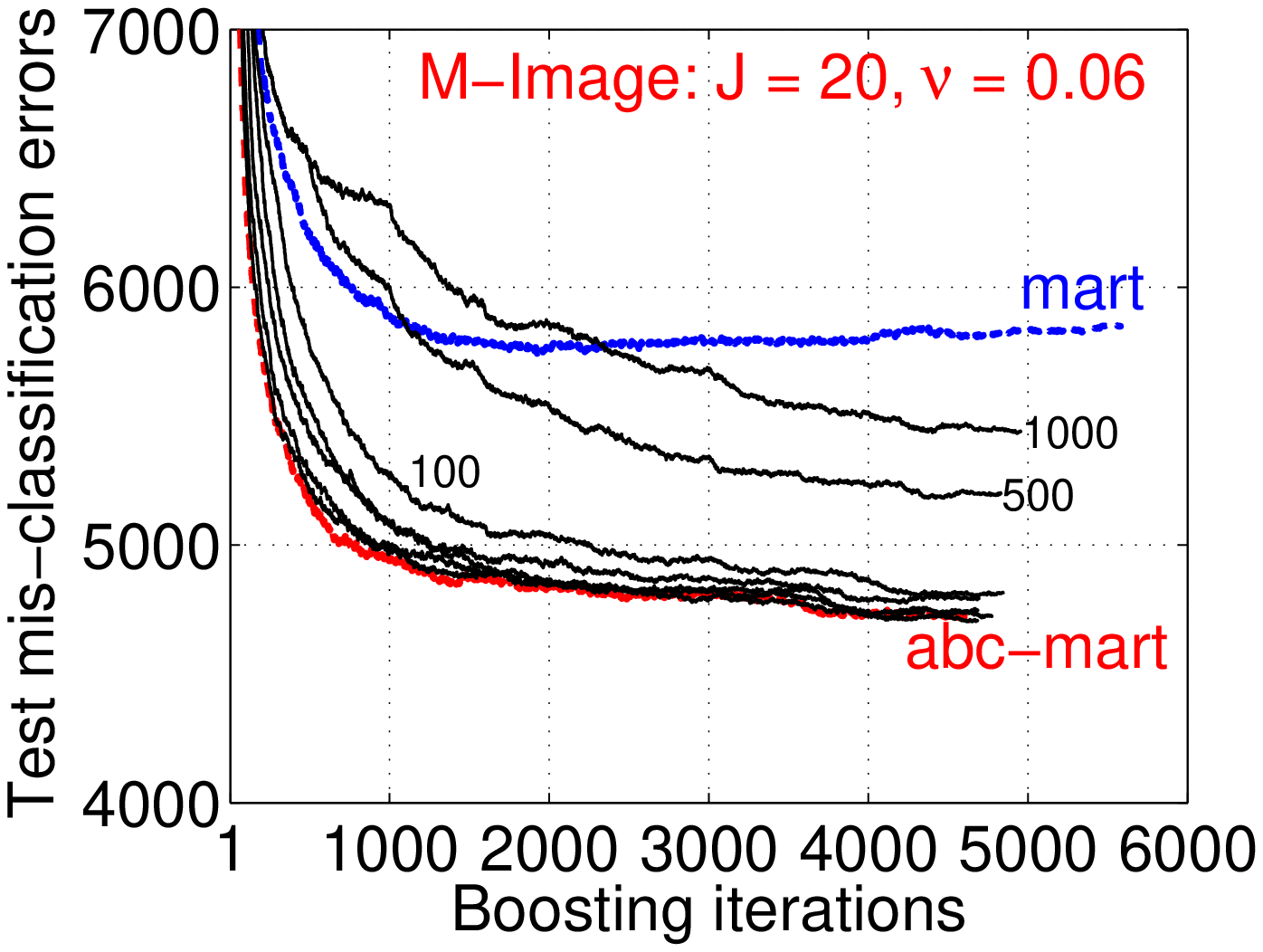}\hspace{-0.1in}
\includegraphics[width = 2.2in]{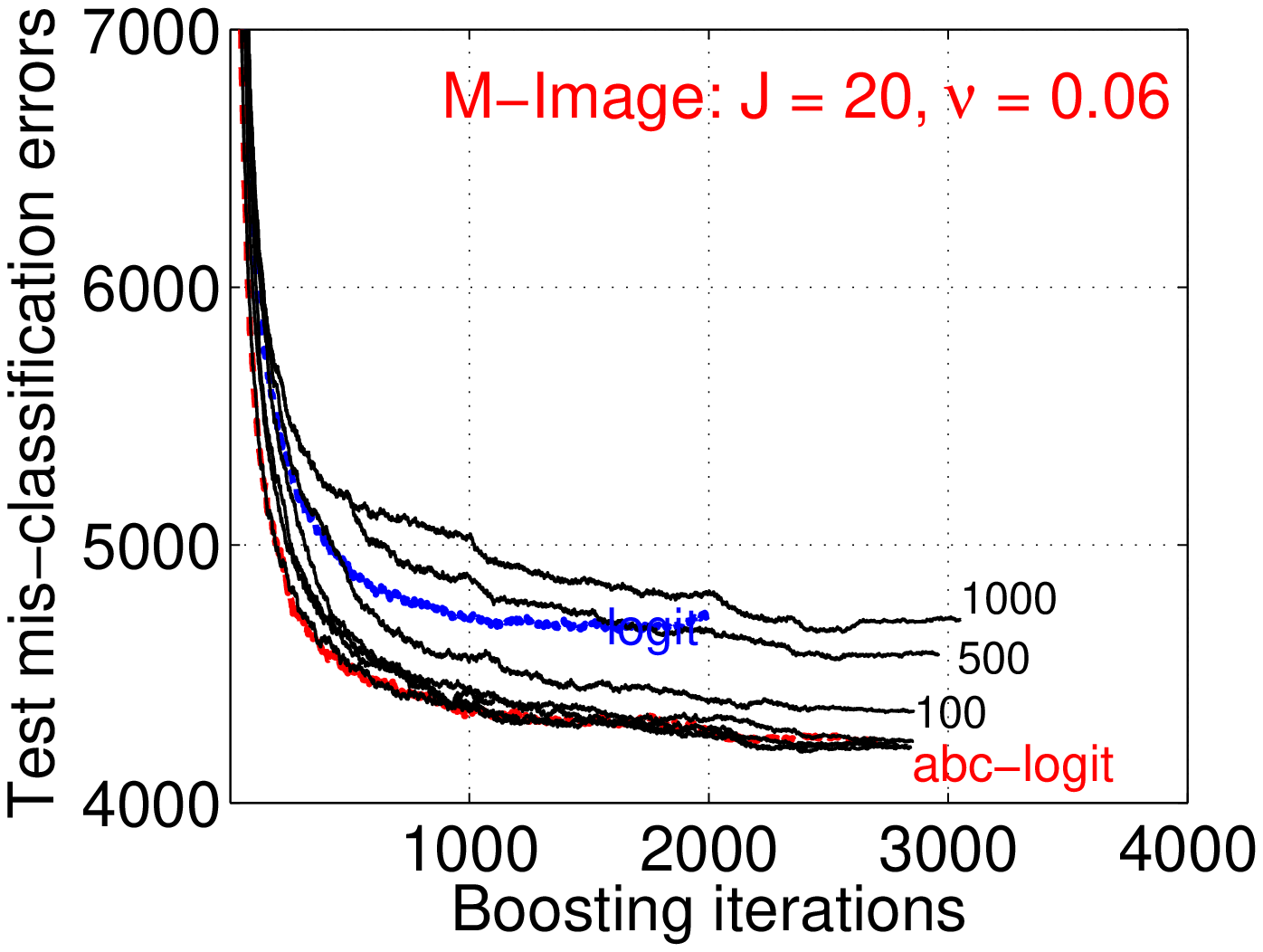}\hspace{-0.1in}
\includegraphics[width = 2.2in]{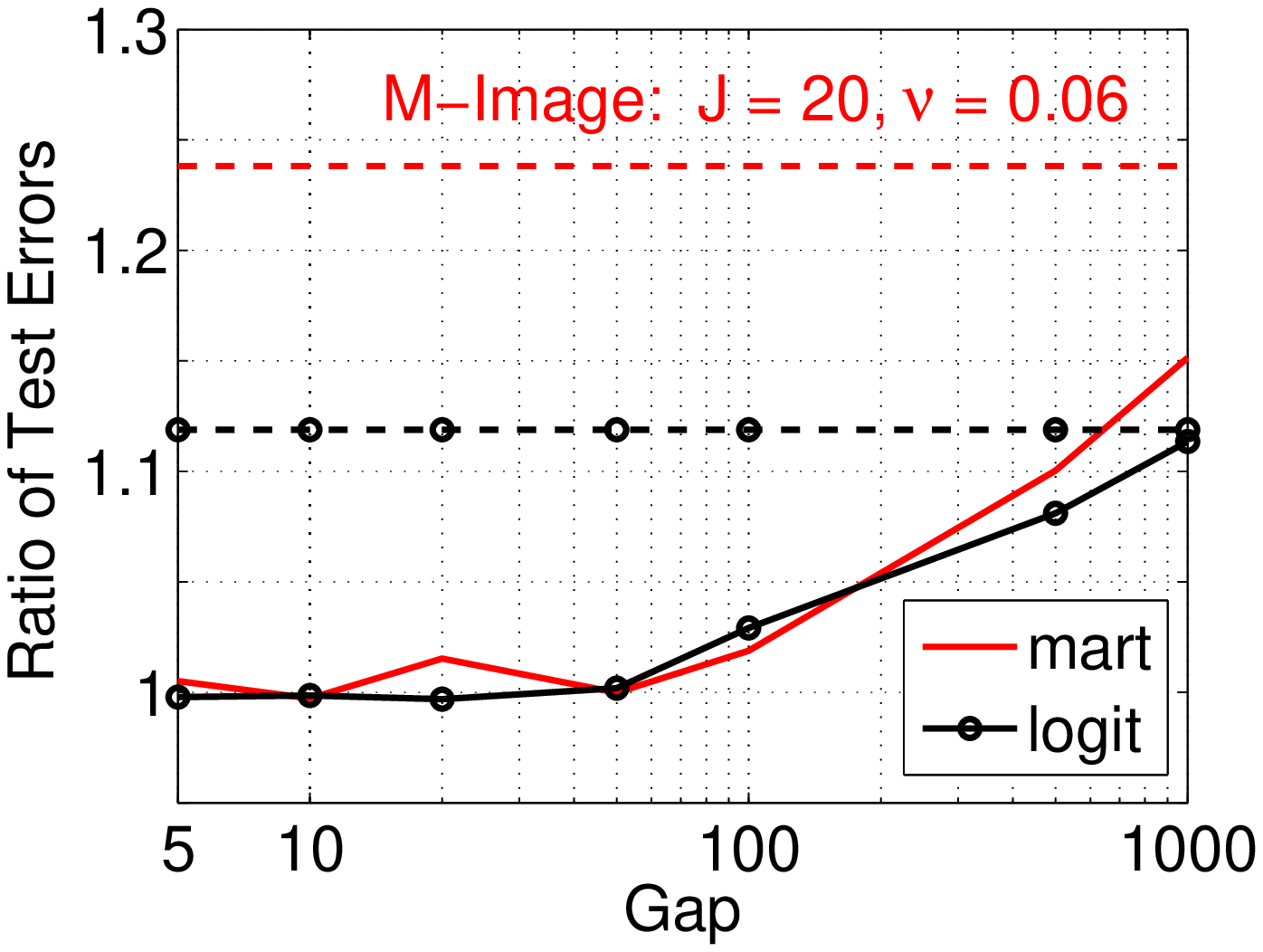}
}

\mbox{
\includegraphics[width = 2.2in]{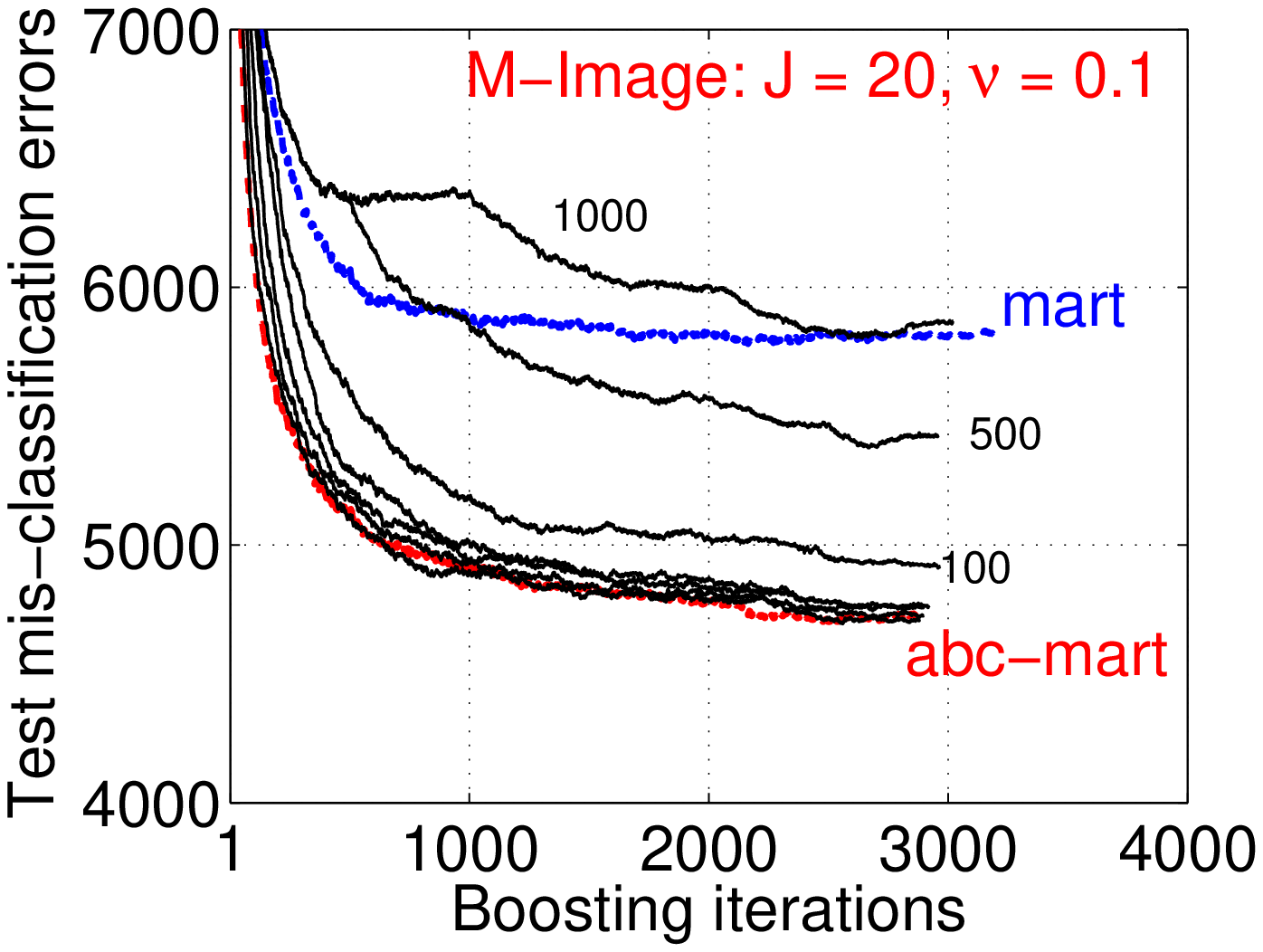}\hspace{-0.1in}
\includegraphics[width = 2.2in]{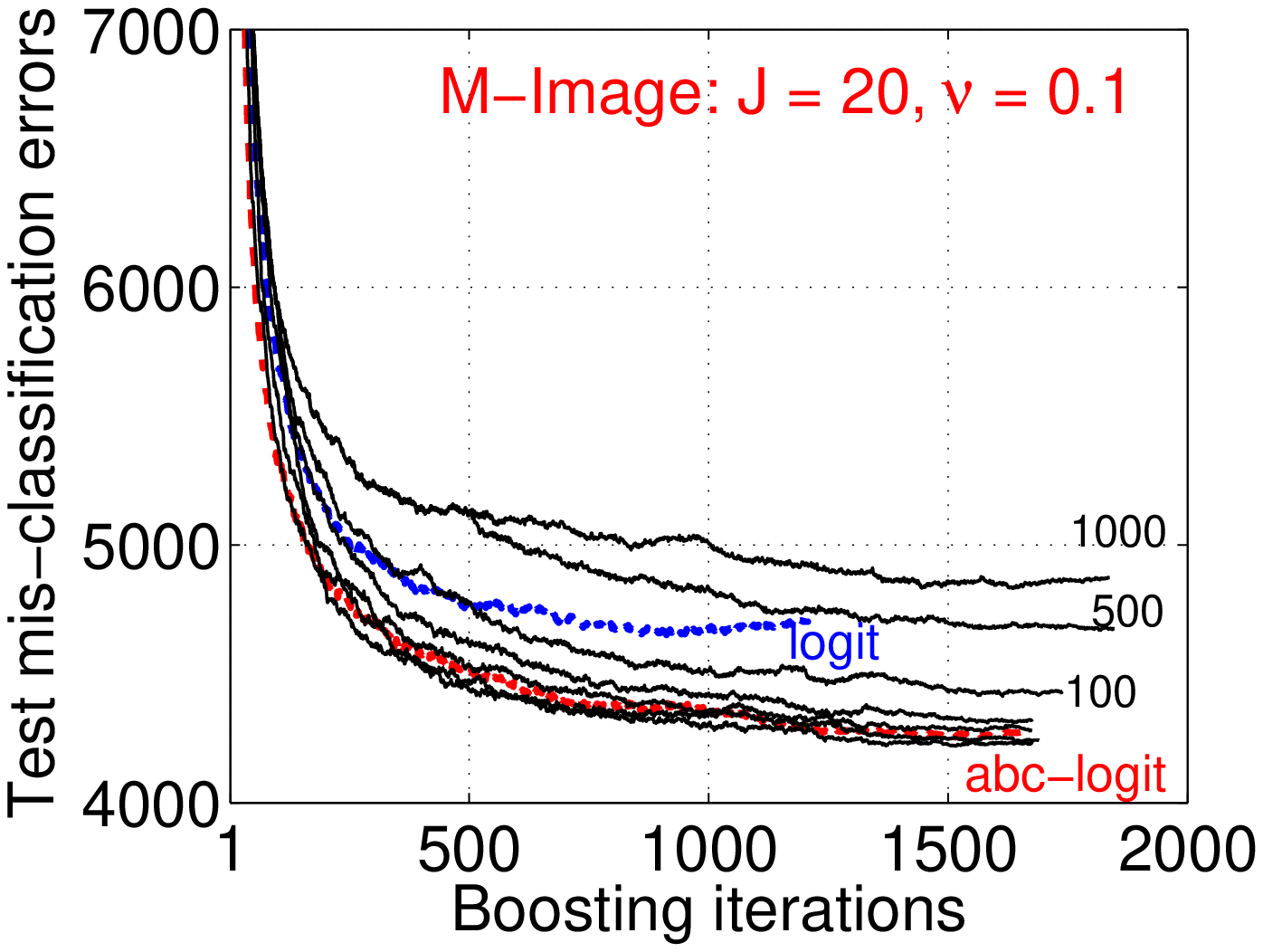}\hspace{-0.1in}
\includegraphics[width = 2.2in]{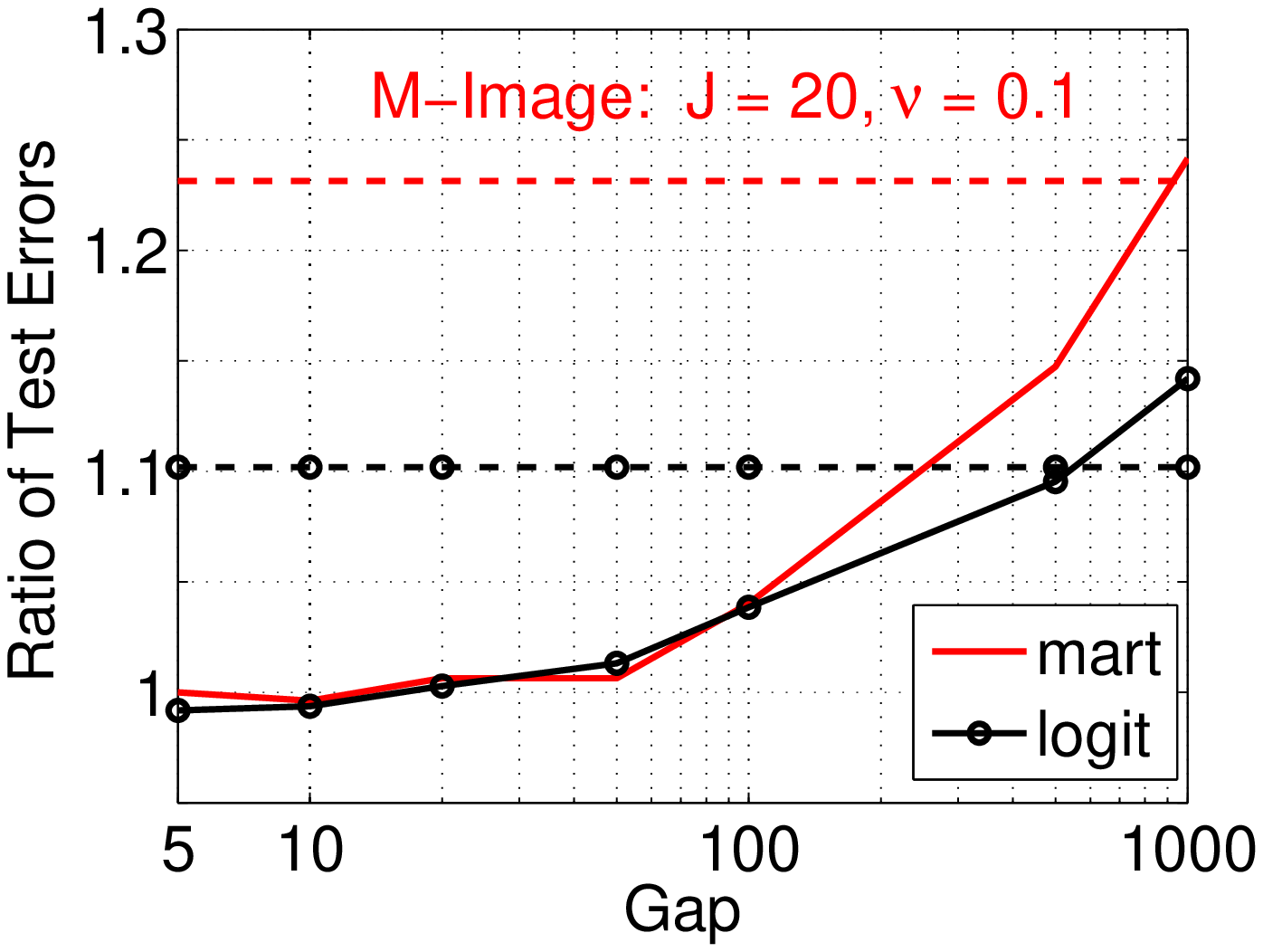}
}
\end{center}\vspace{-0.25in}
\caption{\textbf{M-Image}\hspace{0.2in} See the caption of Figure~\ref{fig_Poker525k} for explanations.}\label{fig_MNIST_imgJ20}
\end{figure}

\begin{figure}[h!]
\begin{center}
\mbox{
\includegraphics[width = 2.2in]{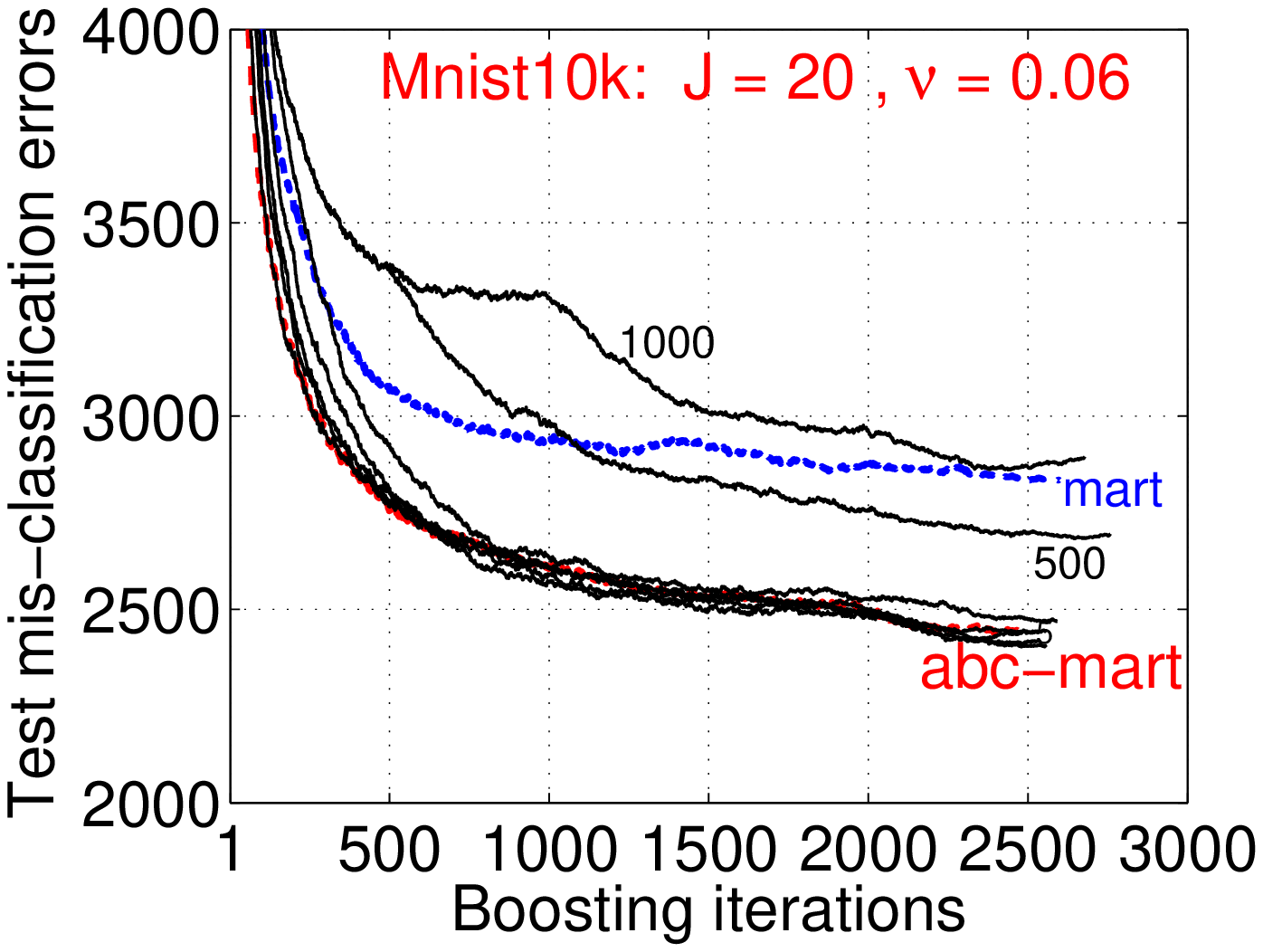}\hspace{-0.1in}
\includegraphics[width = 2.2in]{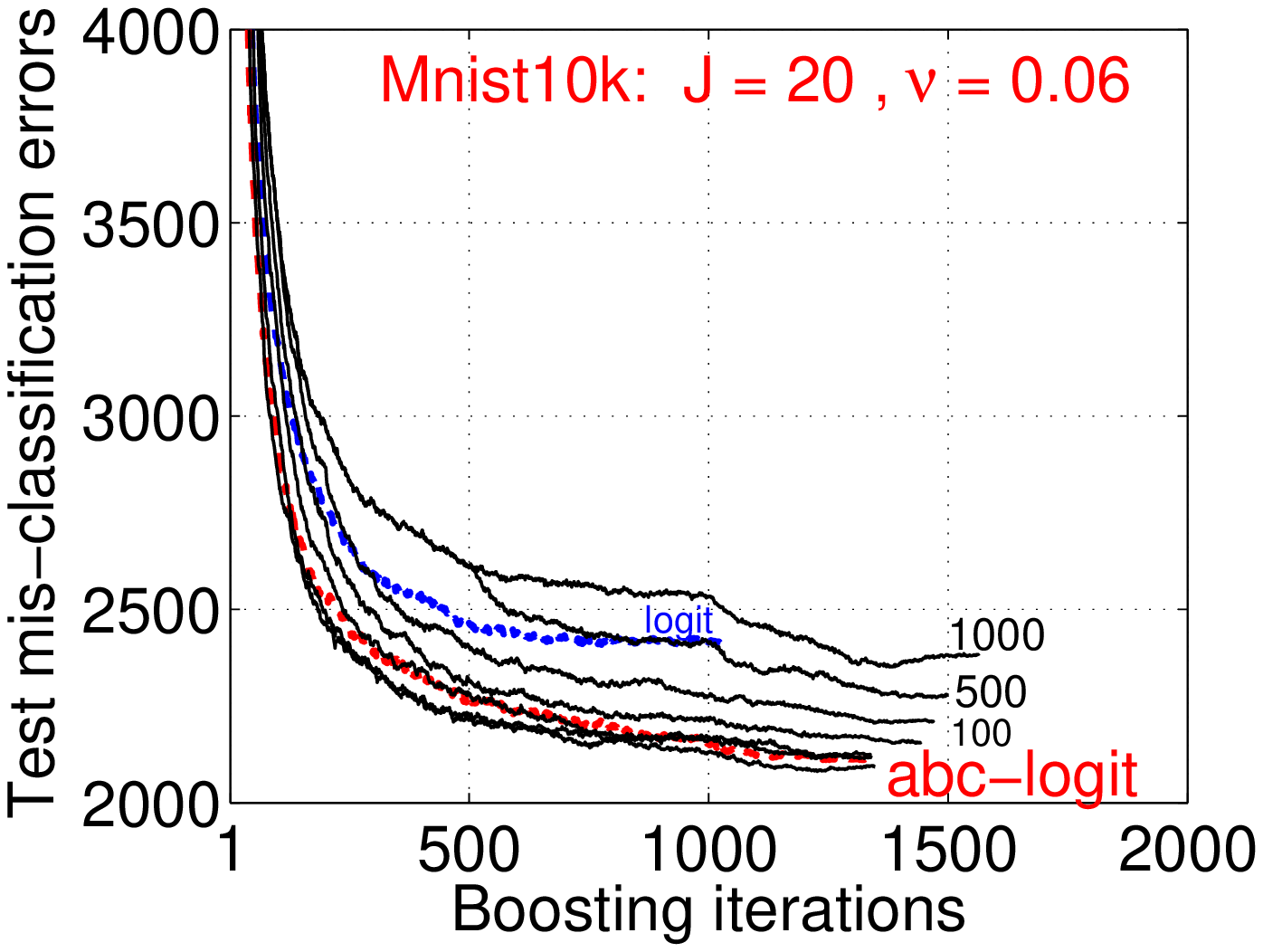}\hspace{-0.1in}
\includegraphics[width = 2.2in]{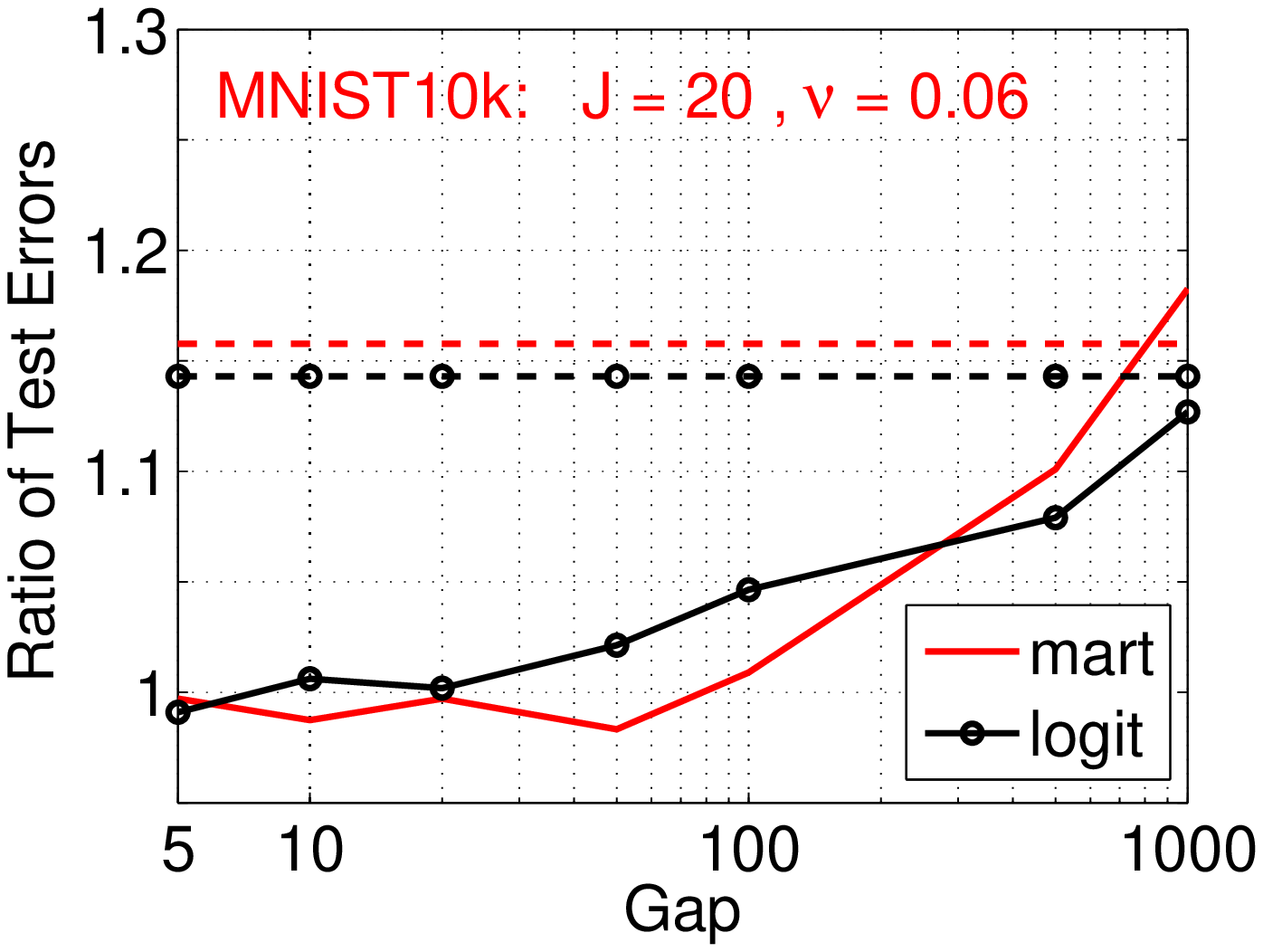}
}

\mbox{
\includegraphics[width = 2.2in]{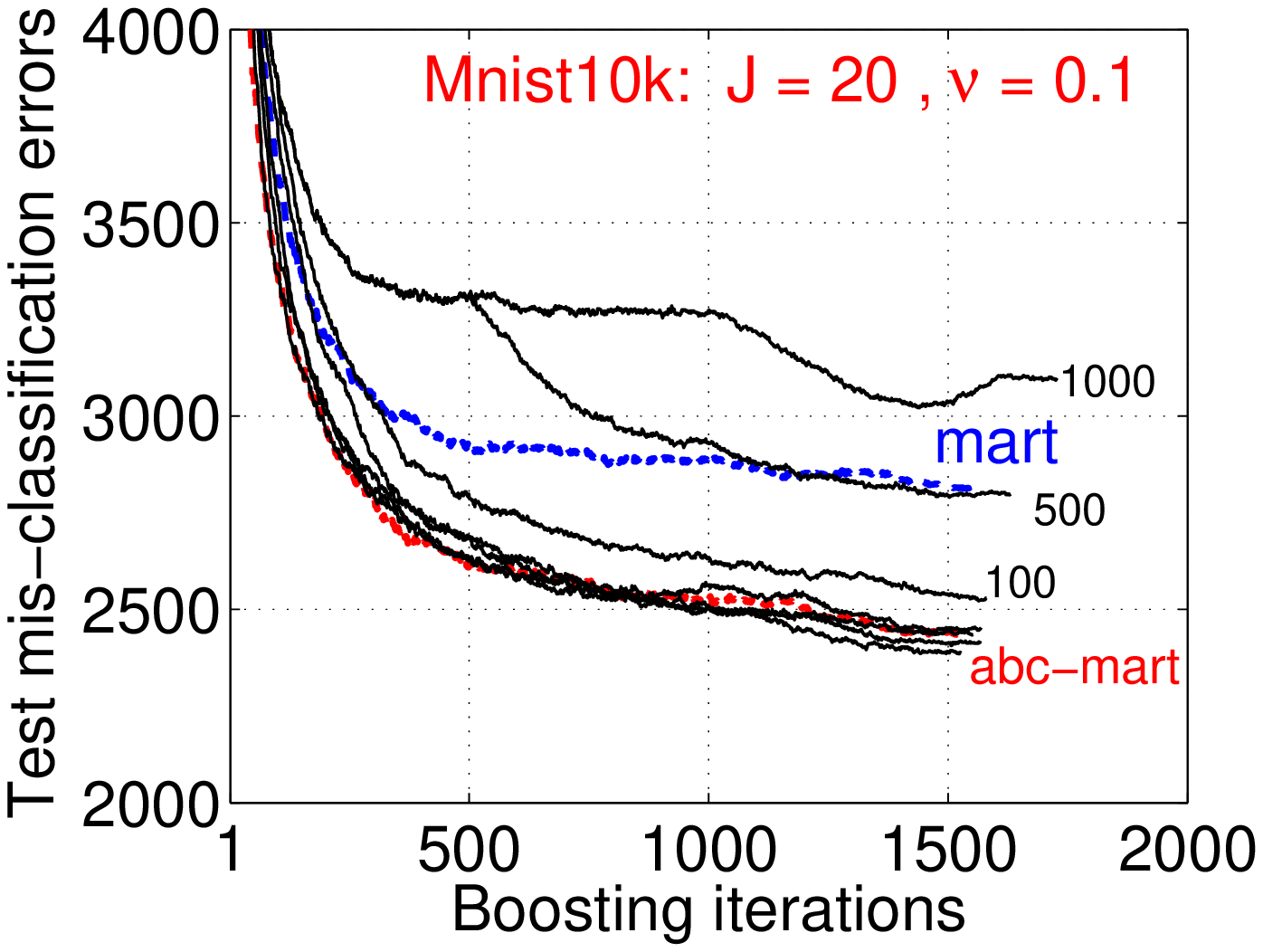}\hspace{-0.1in}
\includegraphics[width = 2.2in]{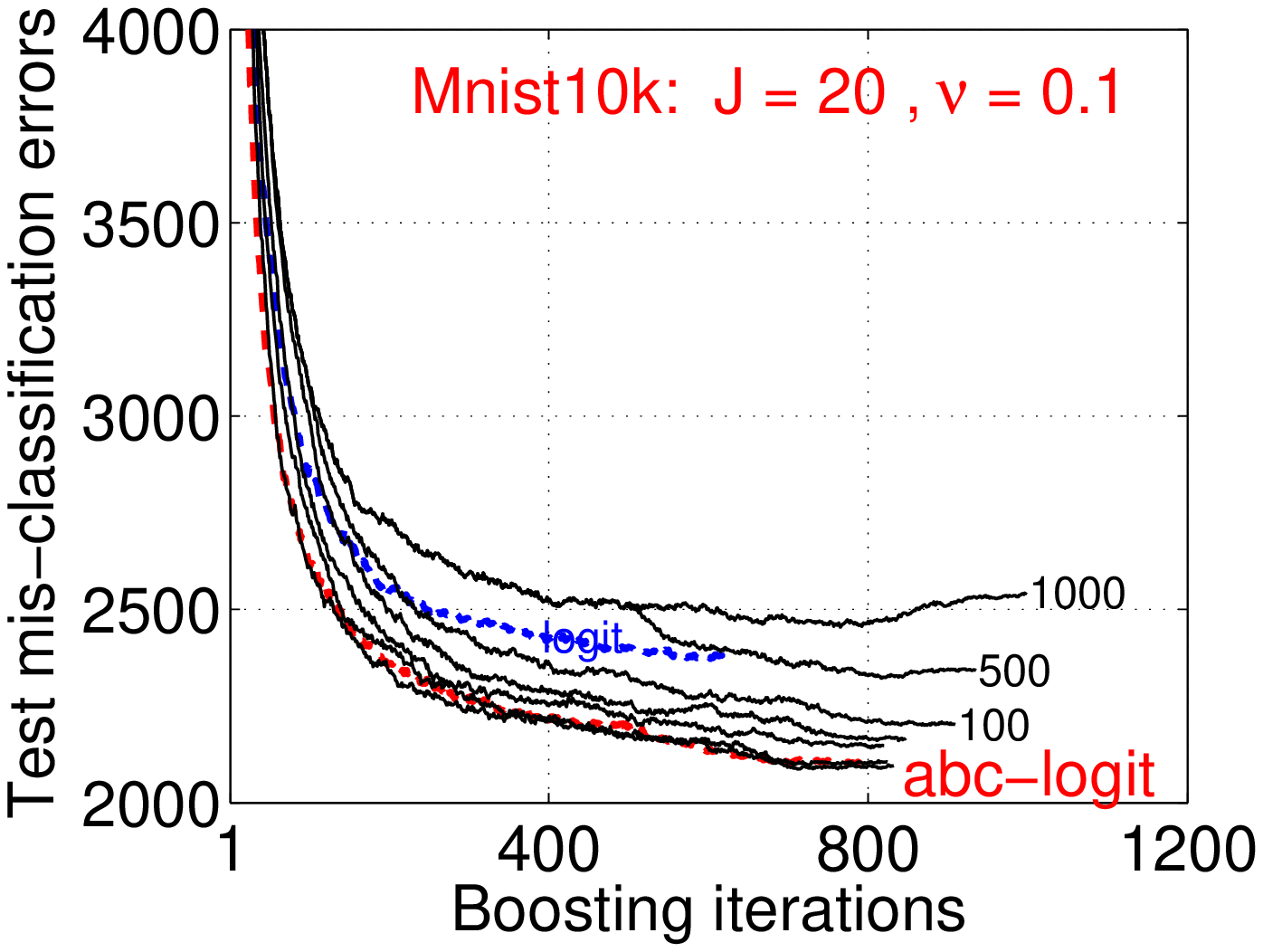}\hspace{-0.1in}
\includegraphics[width = 2.2in]{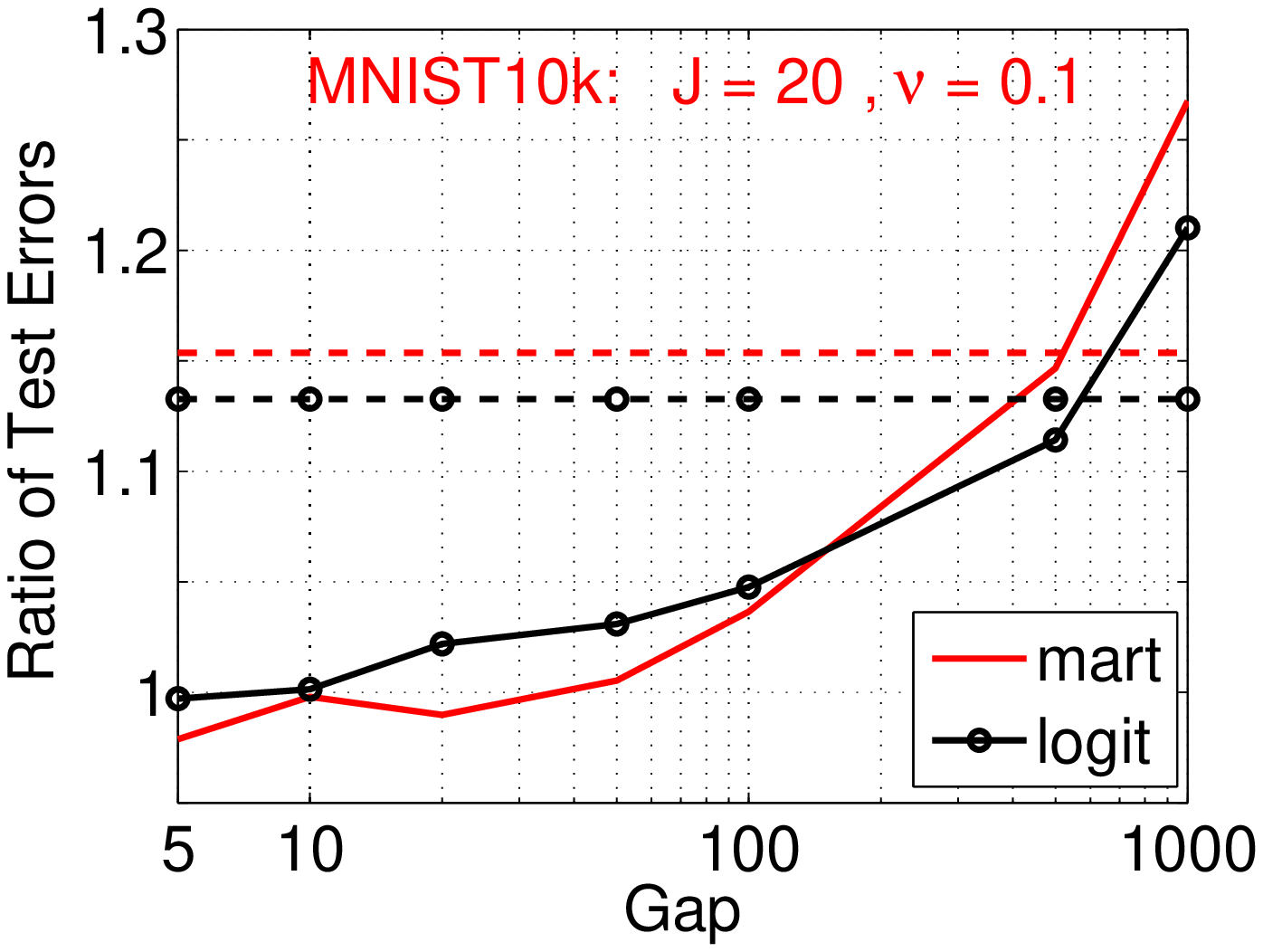}
}
\end{center}\vspace{-0.25in}
\caption{\textbf{Mnist10k}\hspace{0.2in} See the caption of Figure~\ref{fig_Poker525k} for explanations. }\label{fig_MNIST10kJ20}
\end{figure}

Figure~\ref{fig_MNIST_imgJ20} and Figure~\ref{fig_MNIST10kJ20} show that, on {\em M-Image} and {\em Mnist10k}, using {\em fast abc-boost} with $G>50$ can result in non-negligible loss of test accuracies compared to using $G=1$. When $G$ is too large, e.g., $G=1000$, it is possible that {\em fast abc-boost} may produce even larger test errors than {\em mart} or {\em logitboost}.

Figure~\ref{fig_MNIST_imgJ20} and Figure~\ref{fig_MNIST10kJ20} report the test errors for $J=20$ and two shrinkages, $\nu=0.06, 0.1$. It seems that, at the same $G$, using smaller $\nu$ produces slightly better results. 

\clearpage

\vspace{-0.1in}

\vspace{0.1in}

\begin{figure}[h]
\begin{center}

\mbox{
\includegraphics[width = 2.2in]{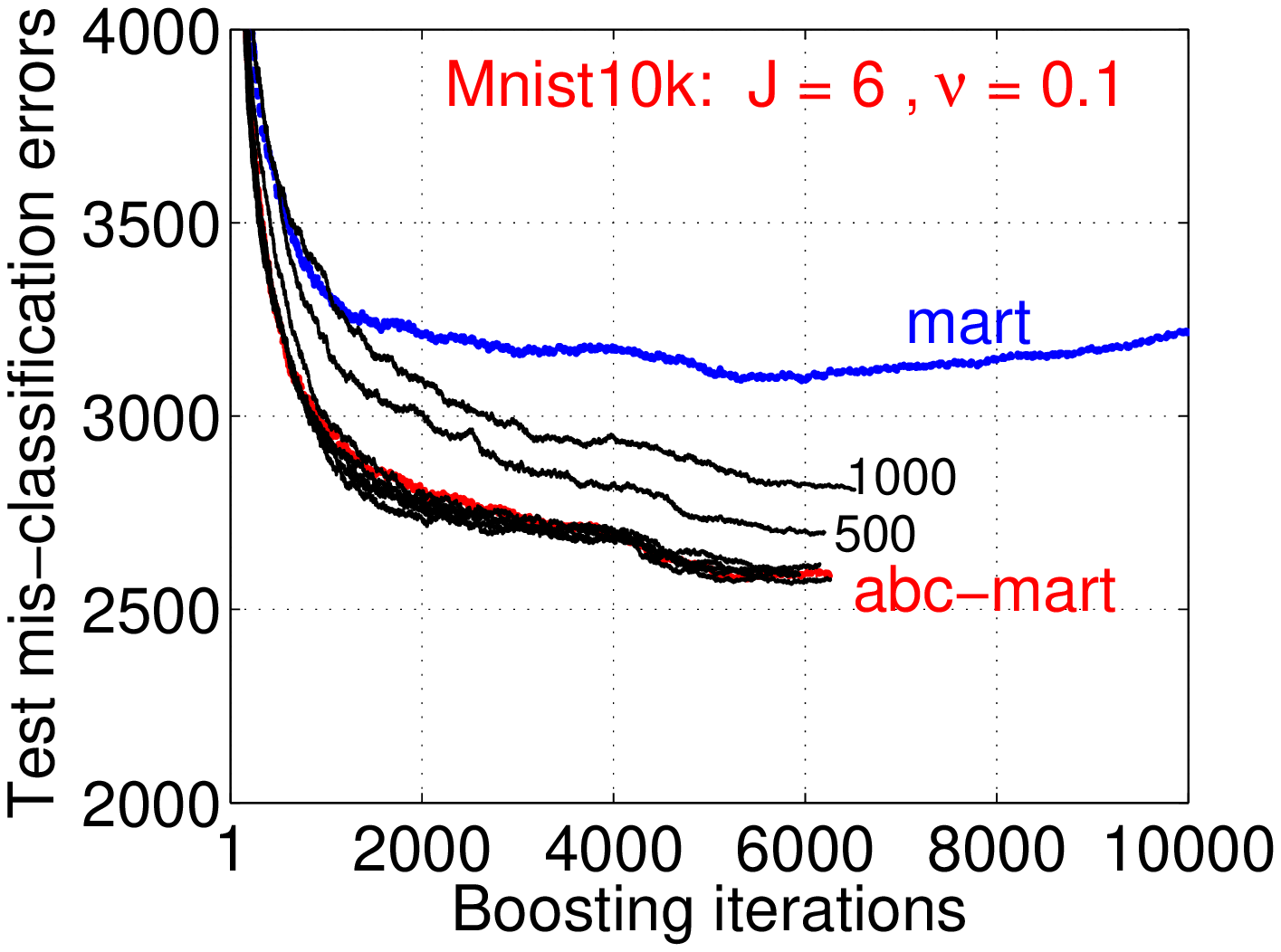}\hspace{-0.1in}
\includegraphics[width = 2.2in]{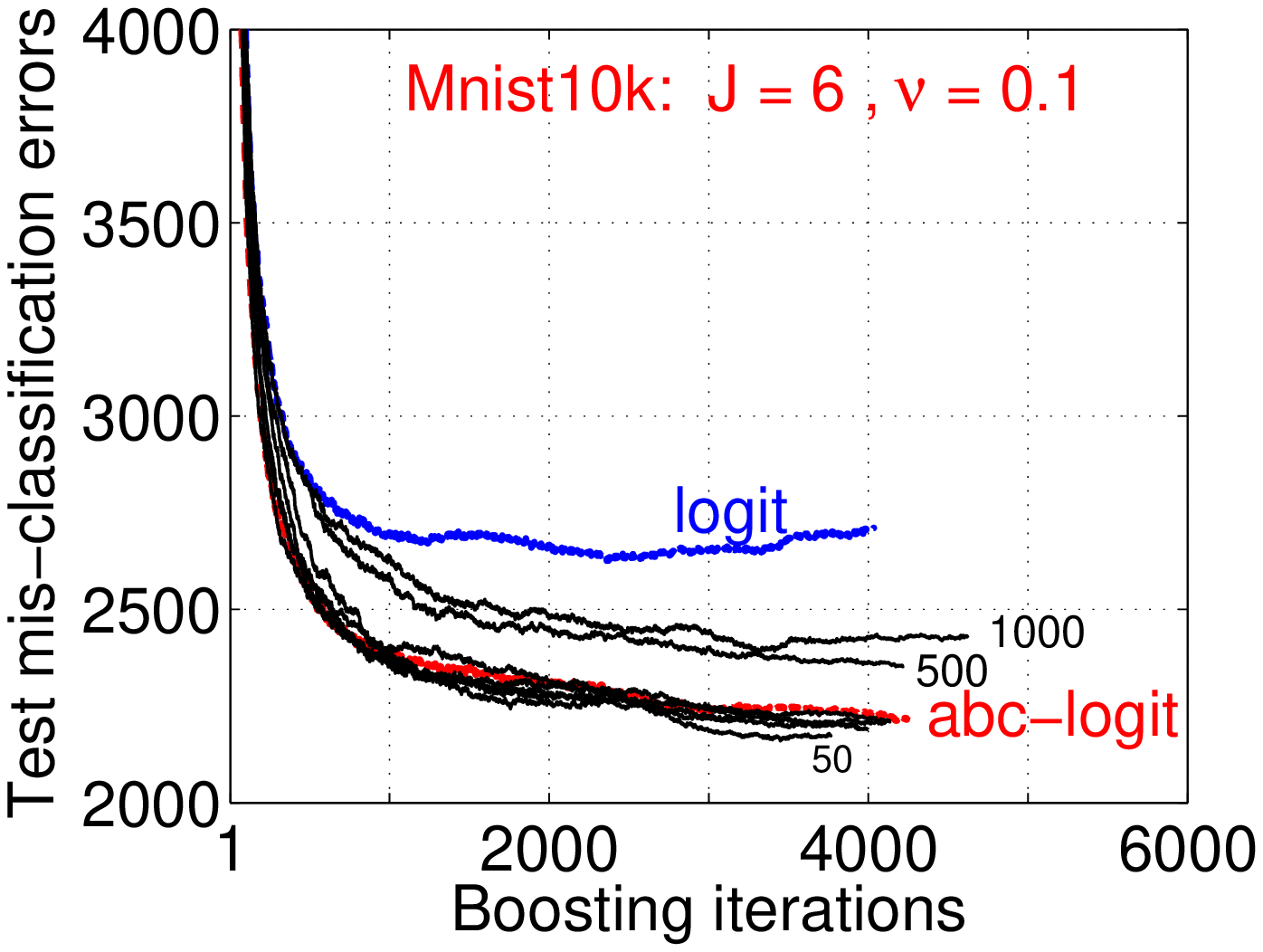}\hspace{-0.1in}
\includegraphics[width = 2.2in]{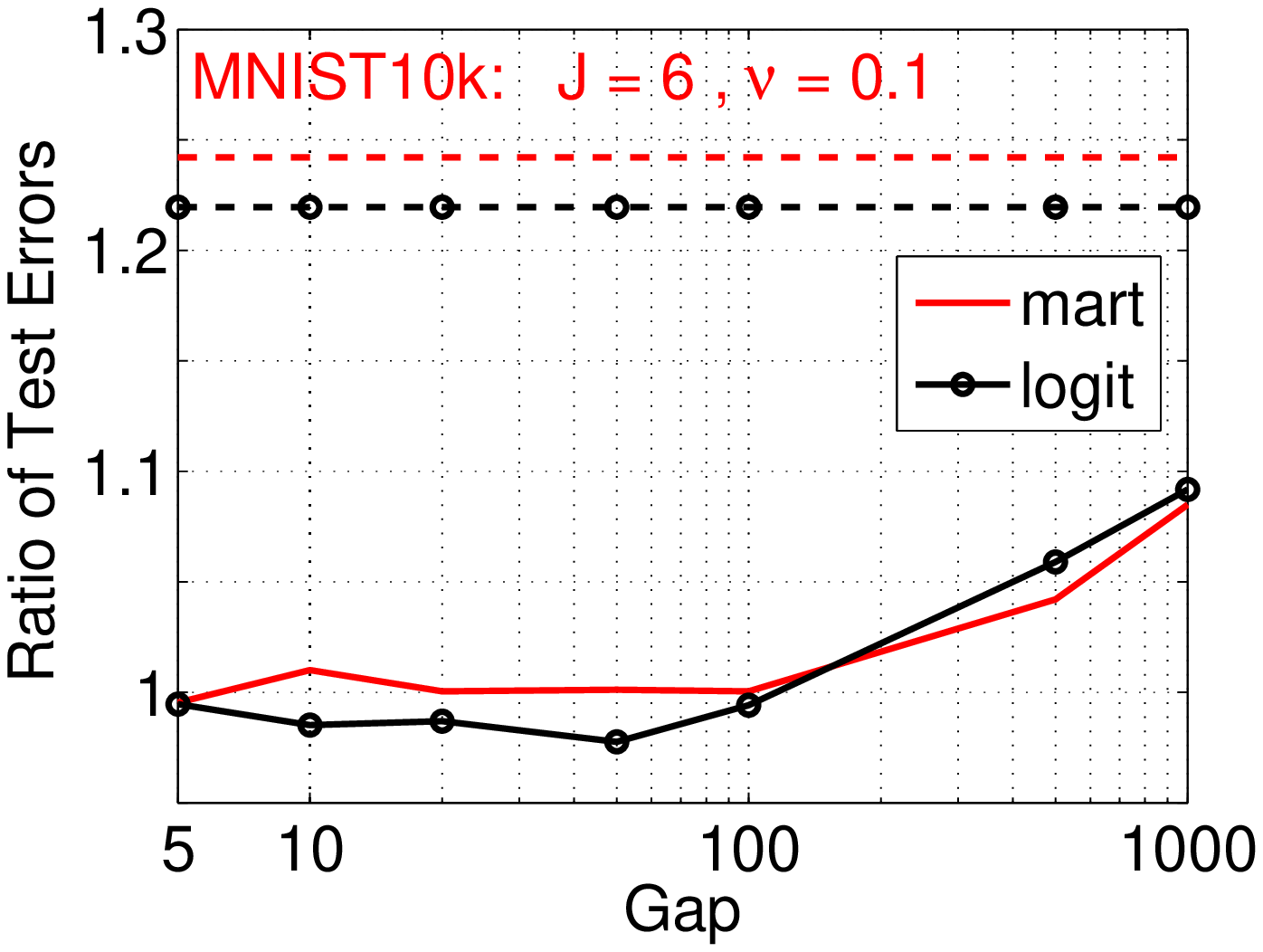}
}

\vspace{-0.12in}

\mbox{
\includegraphics[width = 2.2in]{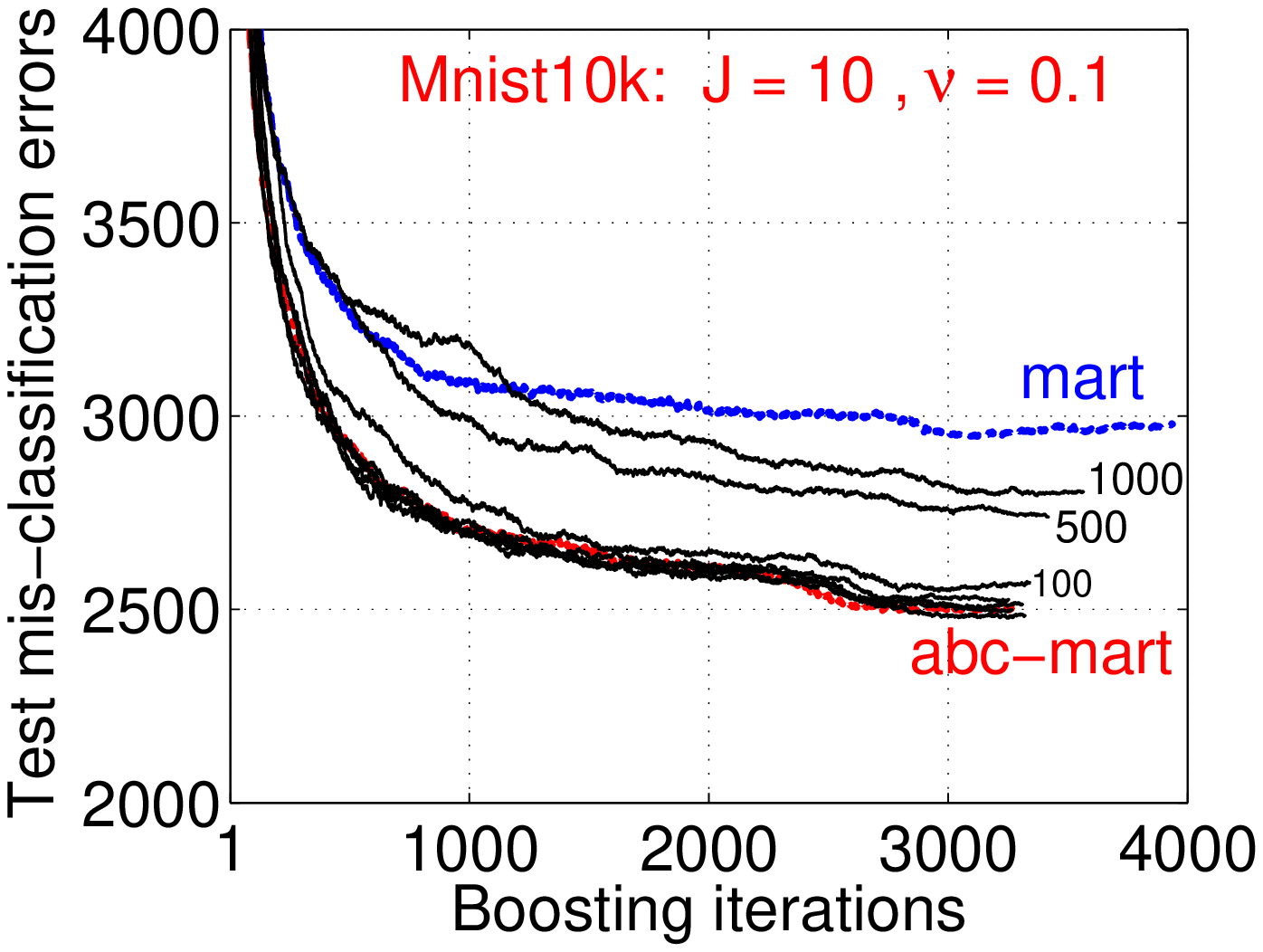}\hspace{-0.1in}
\includegraphics[width = 2.2in]{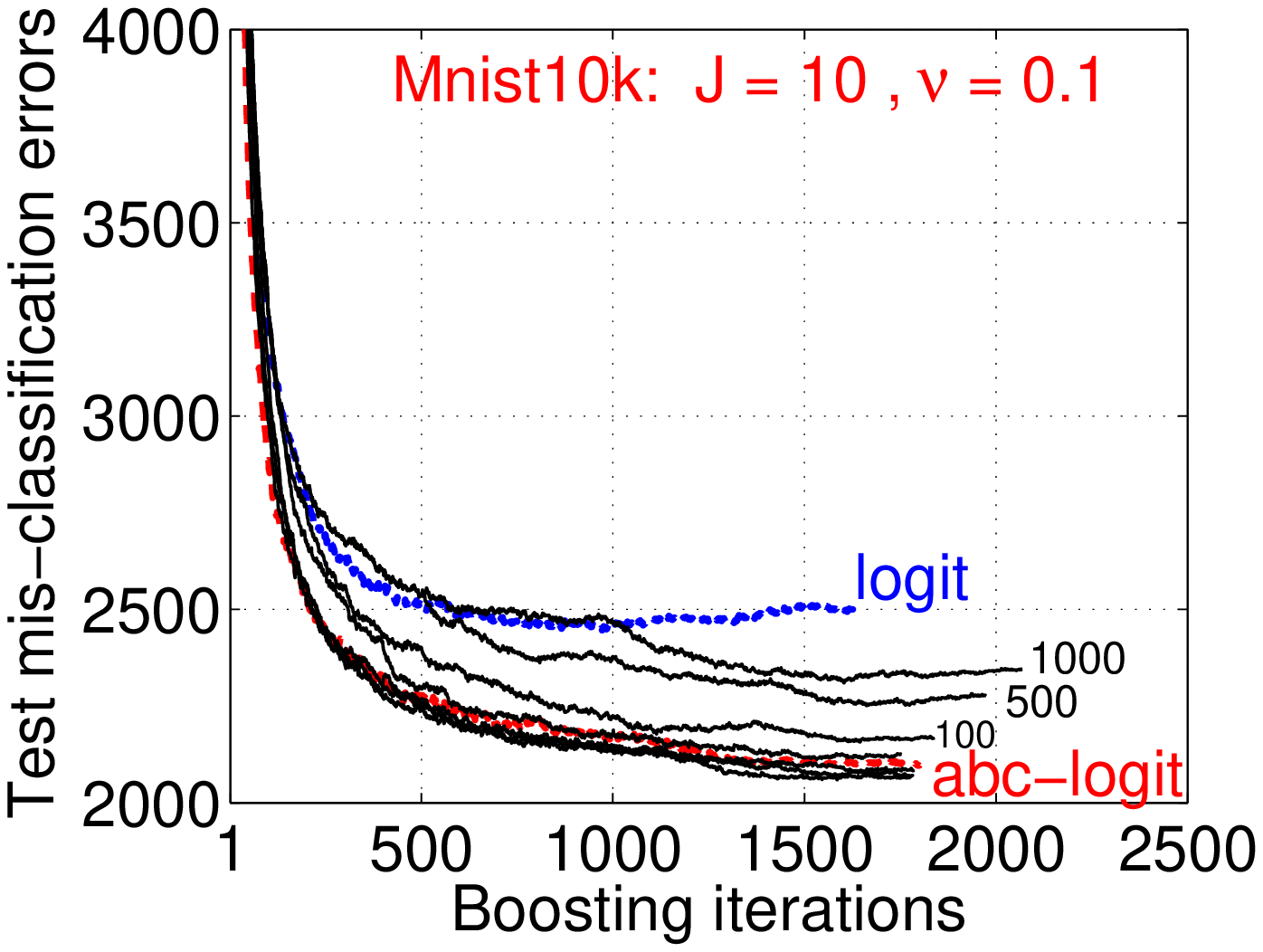}\hspace{-0.1in}
\includegraphics[width = 2.2in]{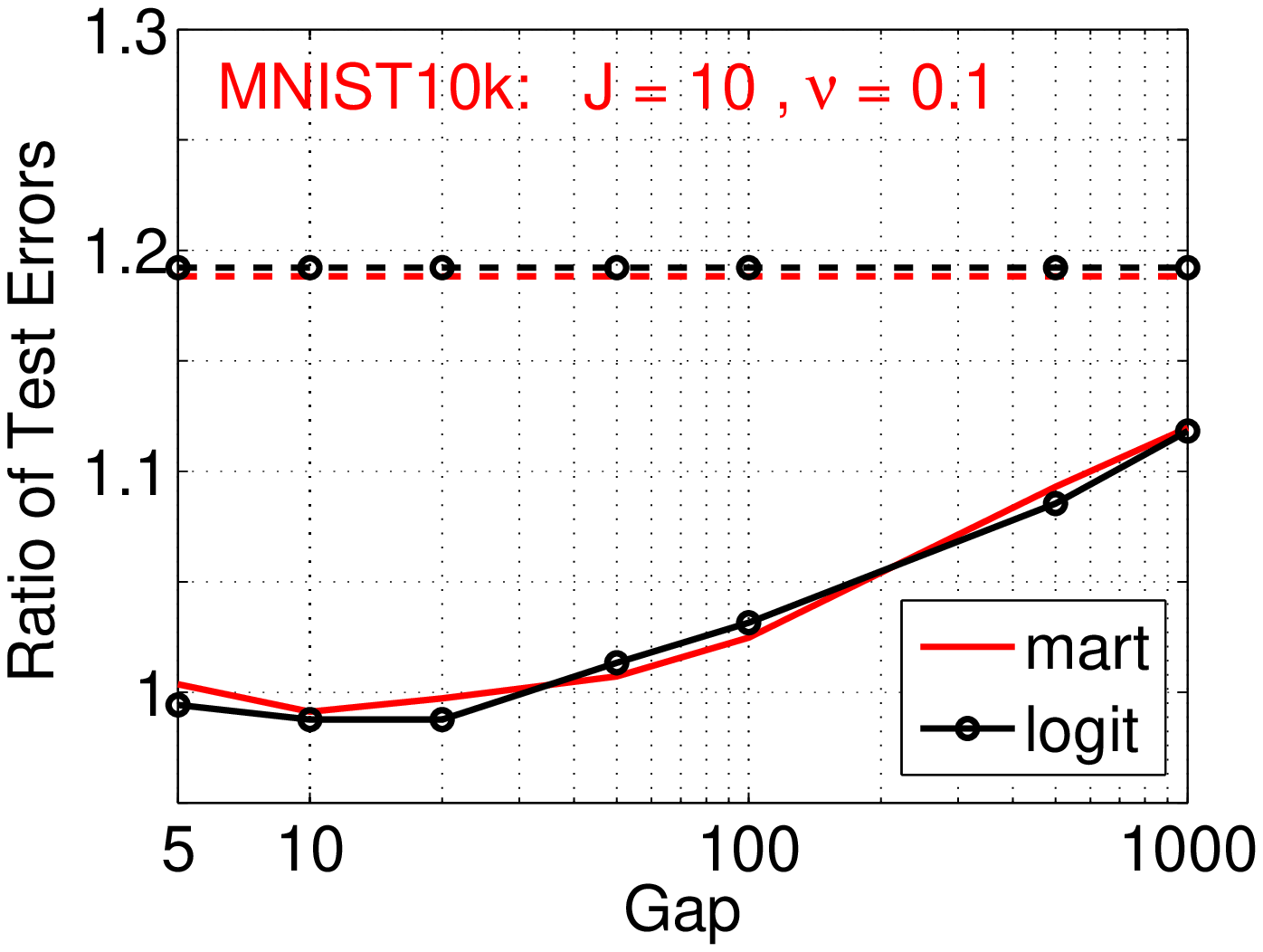}
}

\vspace{-0.12in}

\mbox{
\includegraphics[width = 2.2in]{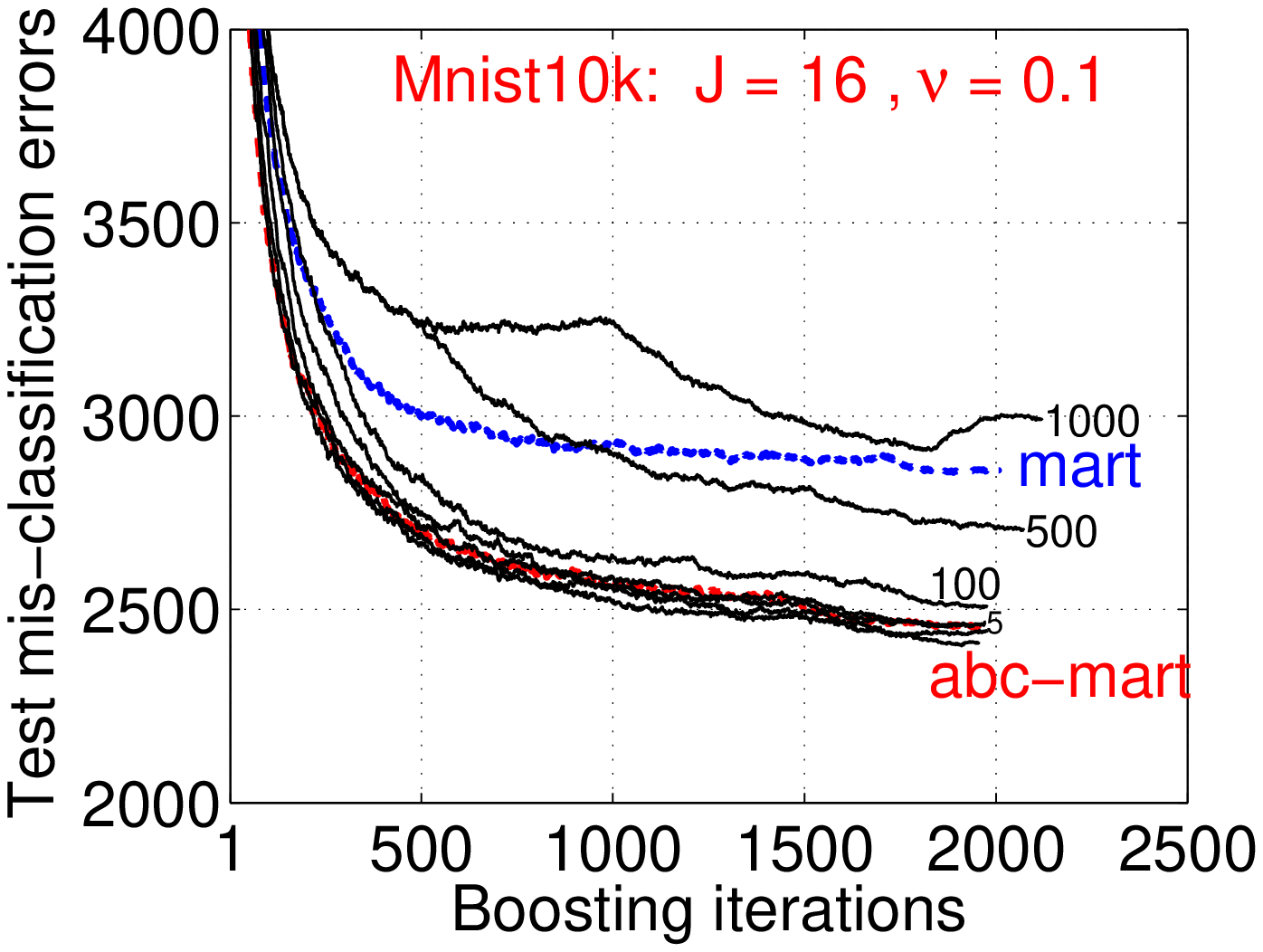}\hspace{-0.1in}
\includegraphics[width = 2.2in]{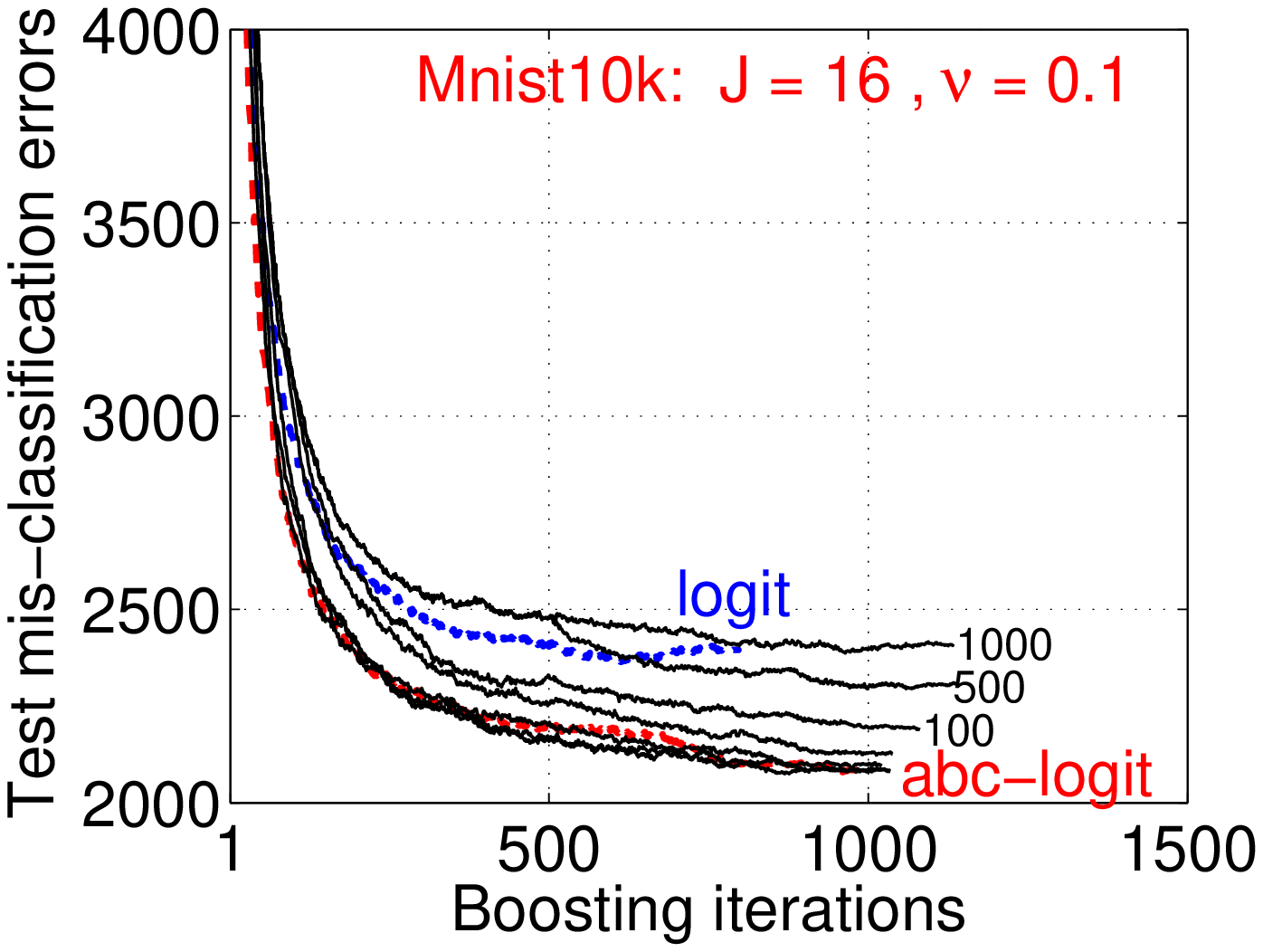}\hspace{-0.1in}
\includegraphics[width = 2.2in]{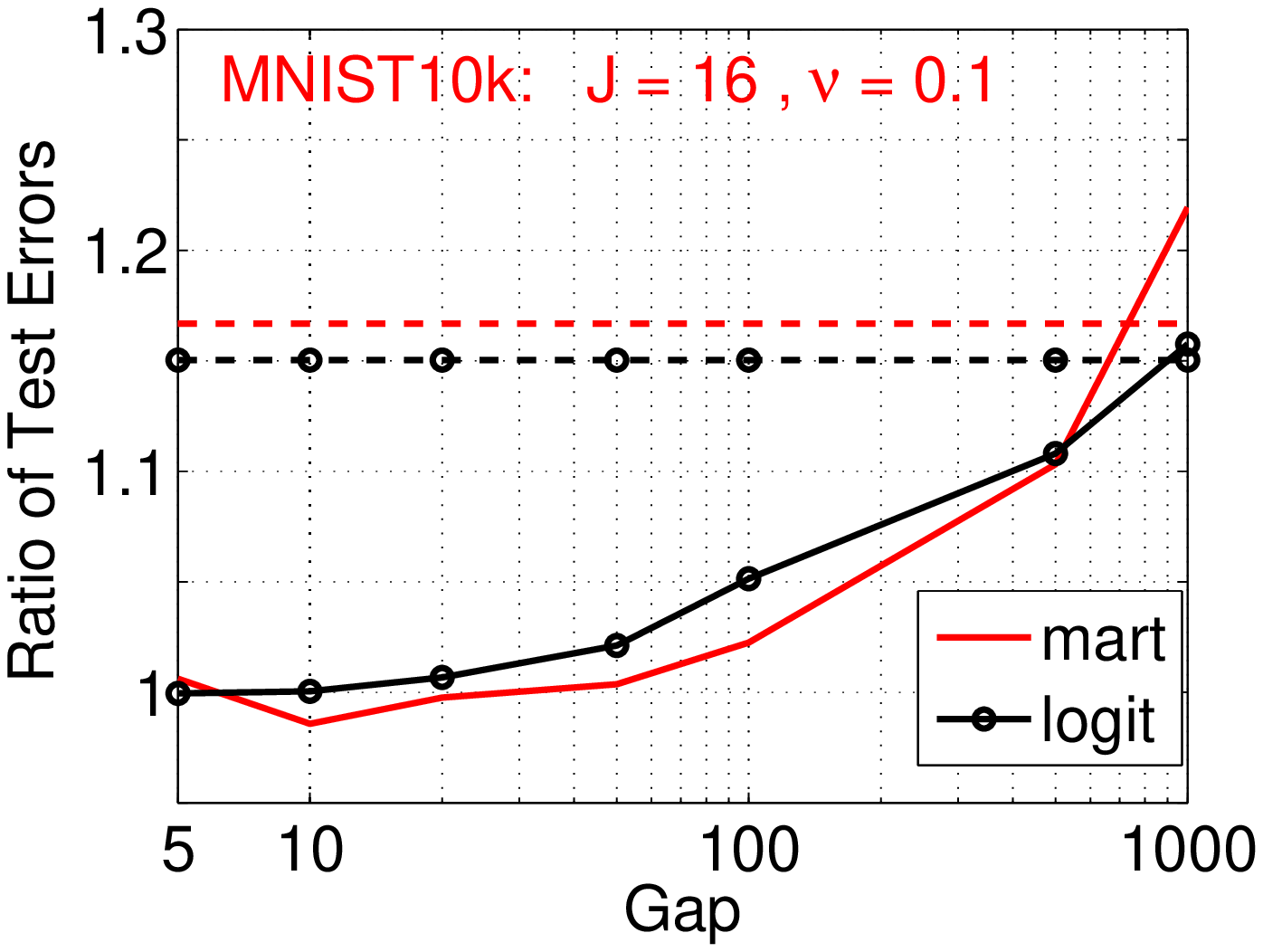}
}

%

\vspace{-0.12in}

\mbox{
\includegraphics[width = 2.2in]{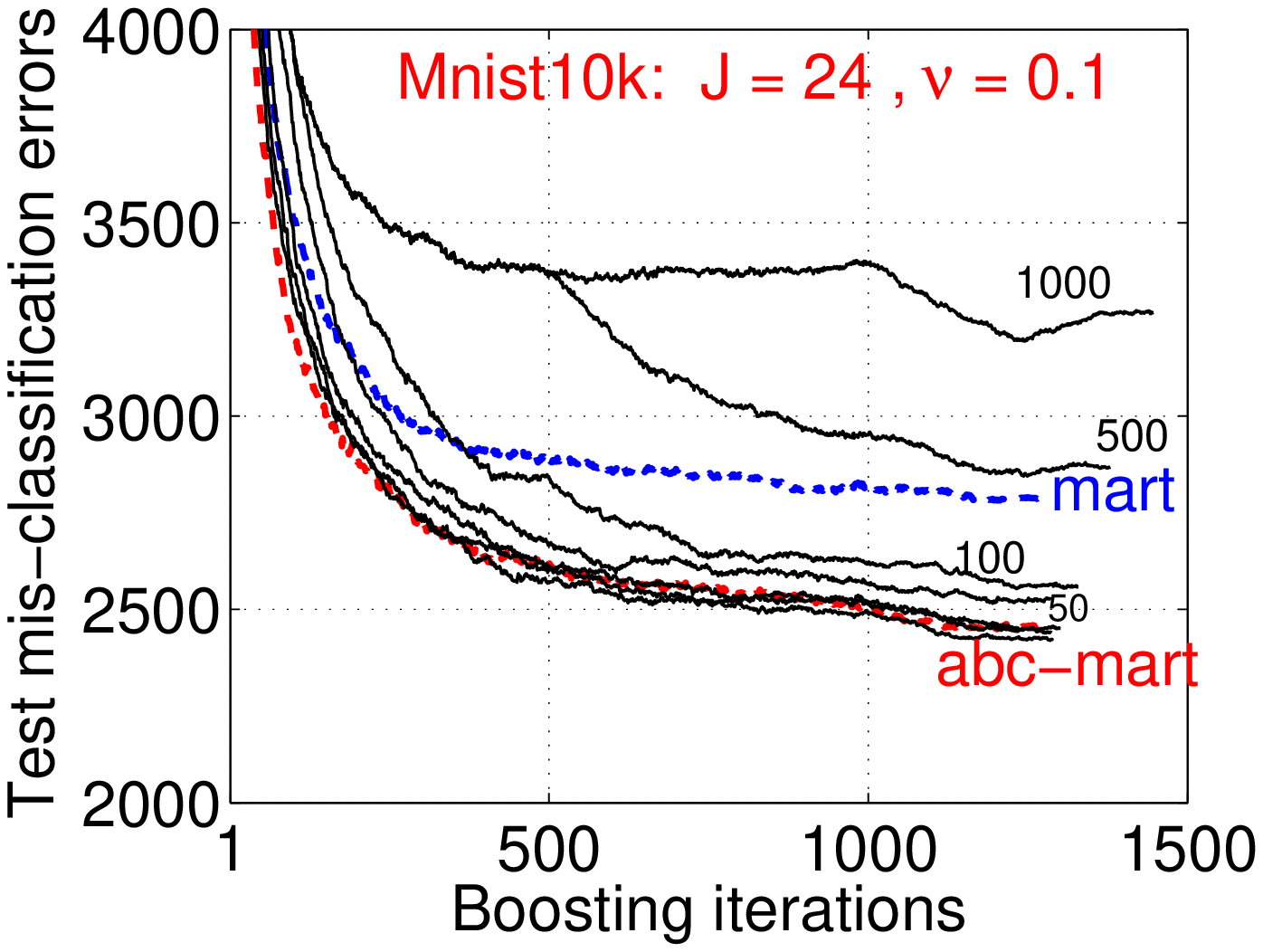}\hspace{-0.1in}
\includegraphics[width = 2.2in]{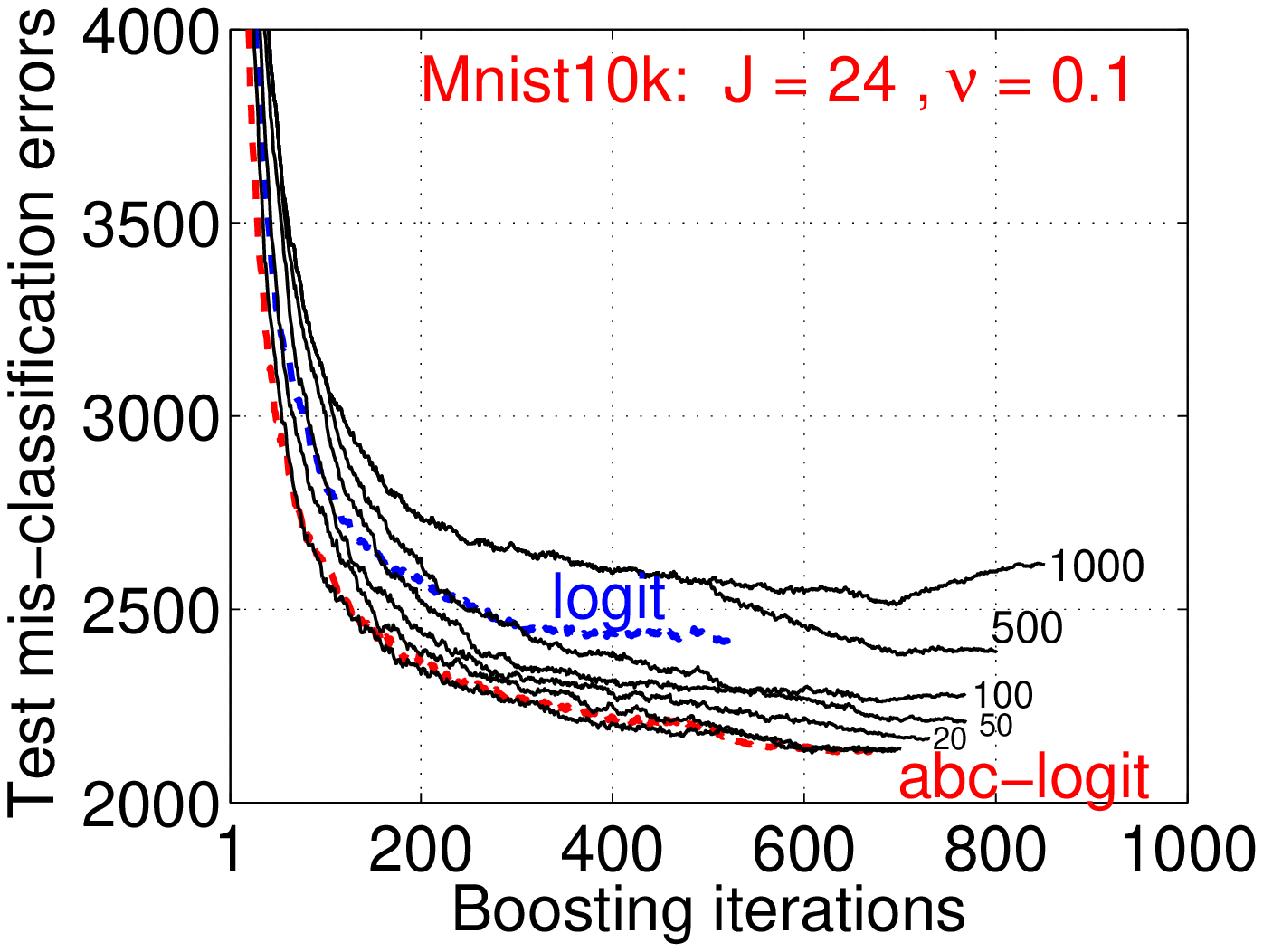}\hspace{-0.1in}
\includegraphics[width = 2.2in]{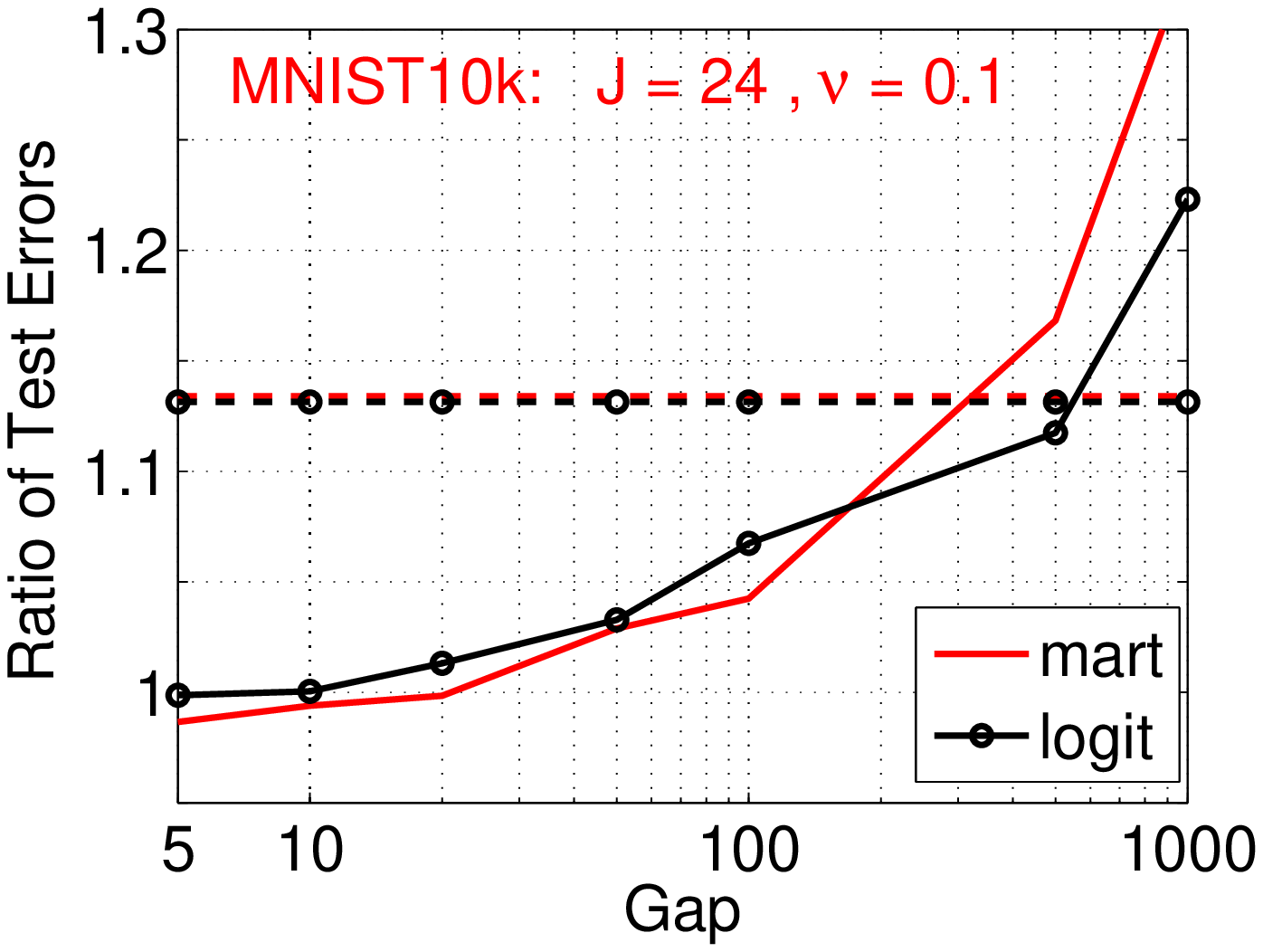}
}

\vspace{-0.12in}

\mbox{
\includegraphics[width = 2.2in]{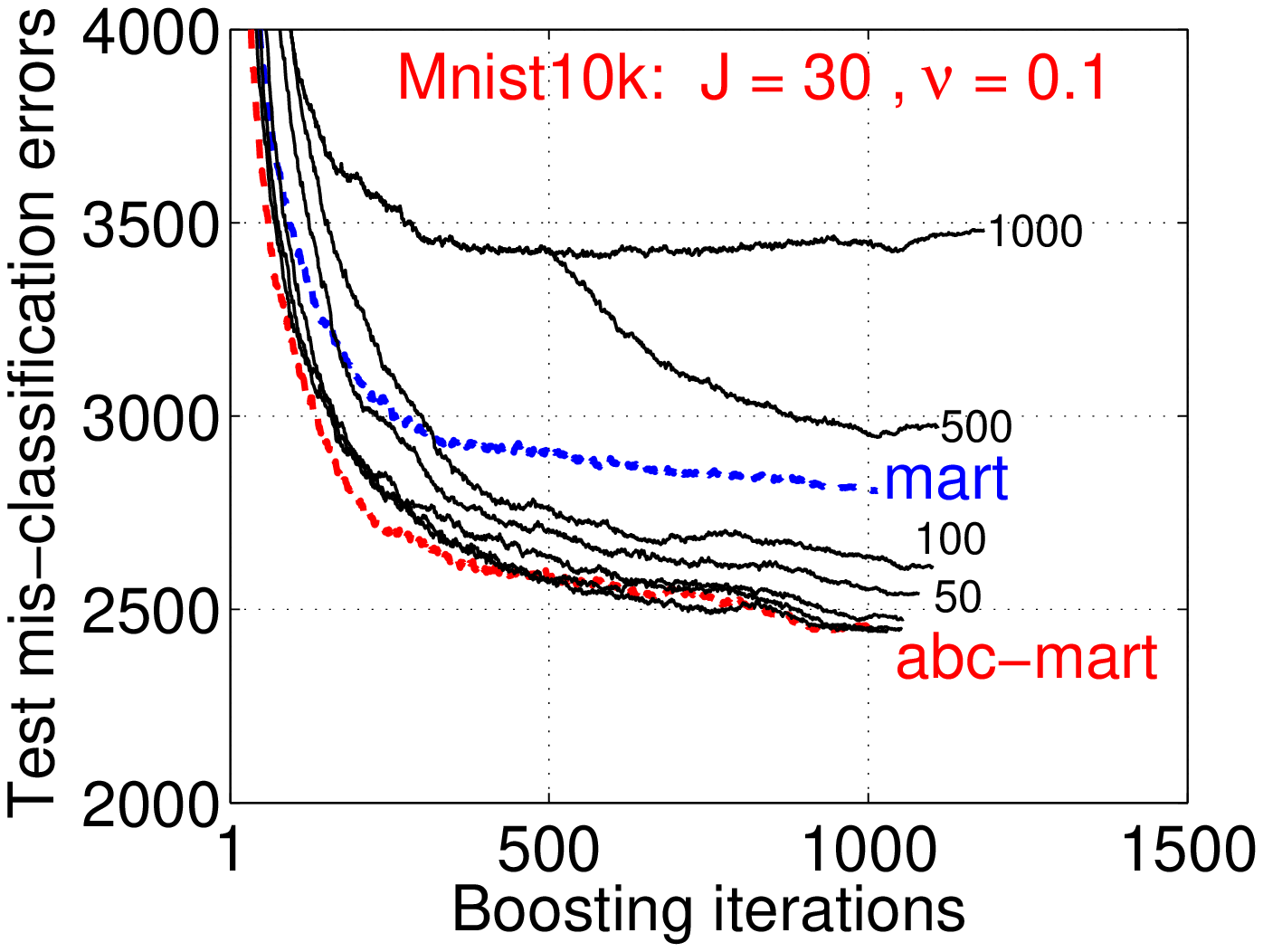}\hspace{-0.1in}
\includegraphics[width = 2.2in]{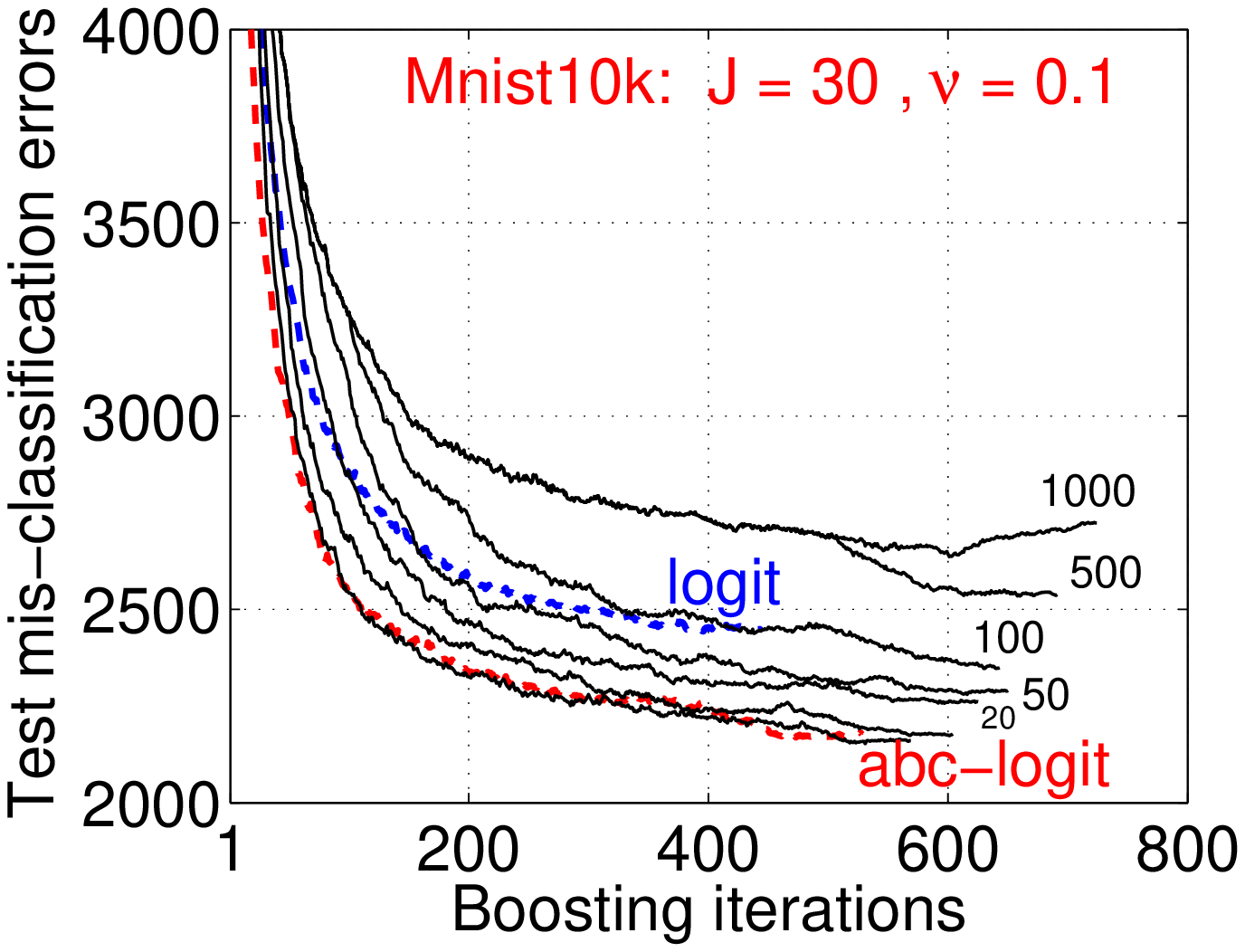}\hspace{-0.1in}
\includegraphics[width = 2.2in]{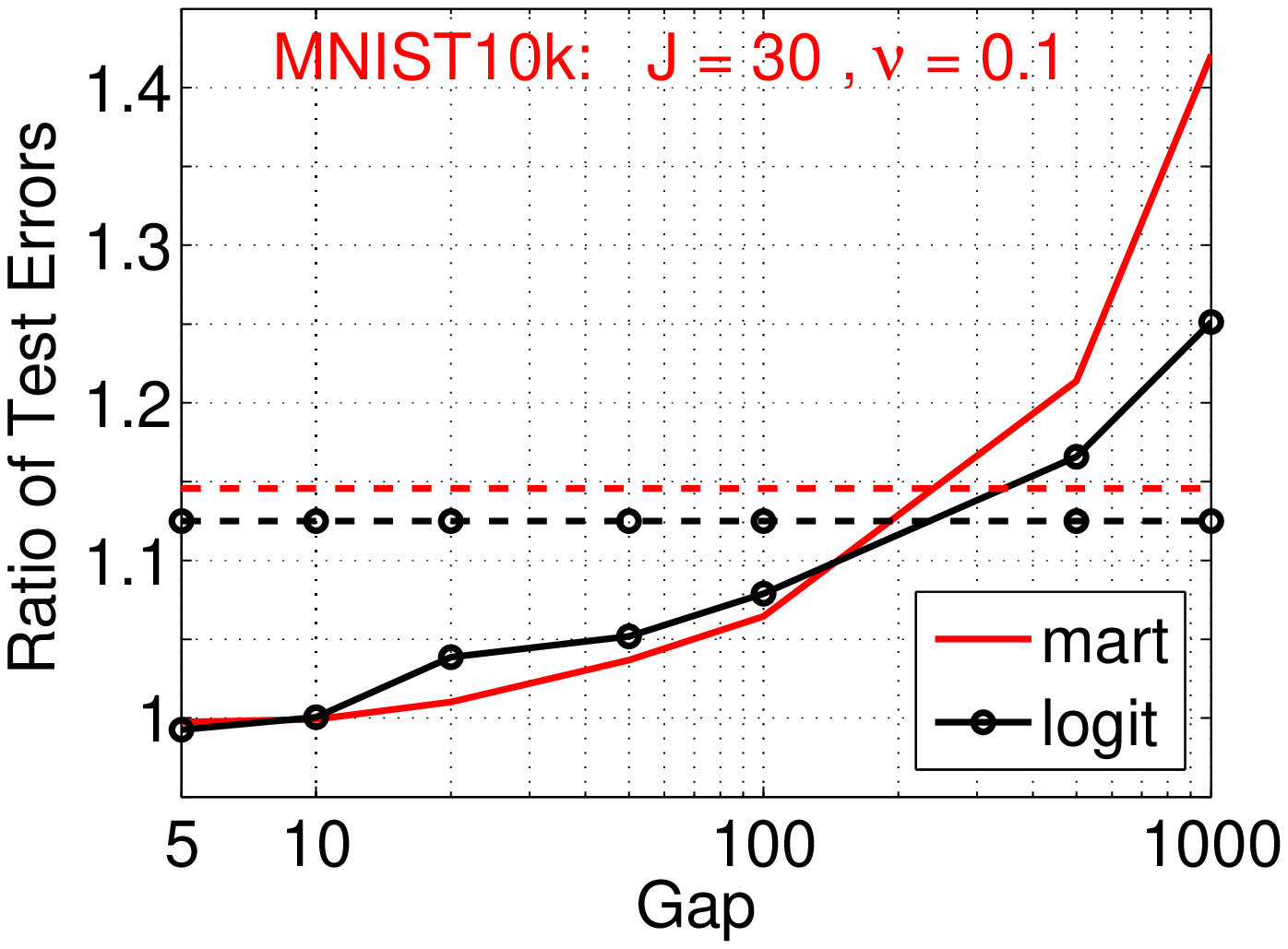}
}

\end{center}\vspace{-0.25in}
\caption{\textbf{Mnist10k}}\label{fig_Mnist10k}
\end{figure}

The above experiments always use $J=20$, which seems to be a reasonable number of terminal tree nodes for large or moderate datasets. Nevertheless, it would be interesting to experiment with other $J$ values.   Figure~\ref{fig_Mnist10k} presents the results on the {\em Mnist10k} dataset, for $J=6, 10, 16, 20, 24, 30$.

When $J$ is small (e.g., $J=6$), using $G$ as large as 100 results in almost no loss of test accuracies. However, when $J$ is large (e.g., $J=30$), even with $G=50$ may produce obviously less accurate results compared to $G=1$.

\clearpage

\section{Conclusion}

This study proposes {\em fast abc-boost} to significantly improve the training speed of {\em abc-boost}, which suffered from serious problems of computational efficiency. {\em Abc-boost} is a new line of boosting algorithms for improving multi-class classification, which was implemented as {\em abc-mart} and {\em abc-logitboost} in prior studies.  {\em Abc-boost} requires that a {\em base class} must be identified at each boosting iteration. The computation of the base class was based on an expensive exhaustive search strategy in prior studies.

With {\em fast abc-boost}, we only need to update the choice of the {\em base class} once for every $G$ iterations, where $G$ can be viewed as {\em Gaps} and used as an additional tuning parameter. Our experiments on fairly large datasets show that the test errors are not sensitive to the choice of $G$, even with $G=100$ or 1000. For datasets of moderate size, our experiments show that, when $G\leq 20\sim 50$, there would be no obvious loss of test accuracies compared to the original {\em abc-boost} algorithms (i.e., $G=1$).

These preliminary results are  very encouraging. We expect {\em fast abc-boost} will be a practical tool for accurate multi-class classification.


\vspace{-0.1in}

{\small
\begin{thebibliography}{10}

\bibitem{Article:Bartlett_AS98}
Peter Bartlett, Yoav Freund, Wee~Sun Lee, and Robert~E. Schapire.
\newblock Boosting the margin: a new explanation for the effectiveness of
  voting methods.
\newblock {\em The Annals of Statistics}, 26(5):1651--1686, 1998.

\bibitem{Article:Buhlmann_JASA03}
Peter B\text{\"{u}}hlmann and Bin Yu.
\newblock Boosting with the \text{L}2 loss: Regression and classification.
\newblock {\em Journal of the American Statistical Association},
  98(462):324--339, 2003.

\bibitem{Article:Freund_95}
Yoav Freund.
\newblock Boosting a weak learning algorithm by majority.
\newblock {\em Inf. Comput.}, 121(2):256--285, 1995.

\bibitem{Article:Freund_JCSS97}
Yoav Freund and Robert~E. Schapire.
\newblock A decision-theoretic generalization of on-line learning and an
  application to boosting.
\newblock {\em J. Comput. Syst. Sci.}, 55(1):119--139, 1997.

\bibitem{Article:Friedman_AS01}
Jerome~H. Friedman.
\newblock Greedy function approximation: A gradient boosting machine.
\newblock {\em The Annals of Statistics}, 29(5):1189--1232, 2001.

\bibitem{Article:FHT_AS00}
Jerome~H. Friedman, Trevor~J. Hastie, and Robert Tibshirani.
\newblock Additive logistic regression: a statistical view of boosting.
\newblock {\em The Annals of Statistics}, 28(2):337--407, 2000.

\bibitem{Article:Lee_JASA04}
Yoonkyung Lee, Yi~Lin, and Grace Wahba.
\newblock Multicategory support vector machines: Theory and application to the
  classification of microarray data and satellite radiance data.
\newblock {\em Journal of the American Statistical Association},
  99(465):67--81, 2004.

\bibitem{Proc:ABC_ICML09}
Ping Li.
\newblock Abc-boost: Adaptive base class boost for multi-class classification.
\newblock In {\em ICML}, pages 625--632, Montreal, Canada, 2009.

\bibitem{Proc:ABC_UAI10}
Ping Li.
\newblock Robust logitboost and adaptive base class (abc) logitboost.
\newblock In {\em UAI}, 2010.

\bibitem{Proc:Mason_NIPS00}
Liew Mason, Jonathan Baxter, Peter Bartlett, and Marcus Frean.
\newblock Boosting algorithms as gradient descent.
\newblock In {\em NIPS}, 2000.

\bibitem{Article:Schapire_ML90}
Robert Schapire.
\newblock The strength of weak learnability.
\newblock {\em Machine Learning}, 5(2):197--227, 1990.

\bibitem{Article:Schapire_ML99}
Robert~E. Schapire and Yoram Singer.
\newblock Improved boosting algorithms using confidence-rated predictions.
\newblock {\em Machine Learning}, 37(3):297--336, 1999.

\bibitem{Article:Tewari_JMLR07}
Ambuj Tewari and Peter~L. Bartlett.
\newblock On the consistency of multiclass classification methods.
\newblock {\em Journal of Machine Learning Research}, 8:1007--1025, 2007.

\bibitem{Article:Zhang_JMLR04}
Tong Zhang.
\newblock Statistical analysis of some multi-category large margin
  classification methods.
\newblock {\em Journal of Machine Learning Research}, 5:1225--1251, 2004.

\bibitem{Article:Zhu_Adaboost09}
Ji~Zhu, Hui Zou, Saharon Rosset, and Trevor Hastie.
\newblock Multi-class adaboost.
\newblock {\em Statistics and Its Interface}, 2(3):349--360, 2009.

\bibitem{Article:Zou_AOAS08}
Hui Zou, Ji~Zhu, and Trevor Hastie.
\newblock New multicategory boosting algorithms based on multicategory
  fisher-consistent losses.
\newblock {\em The Annals of Applied Statistics}, 2(4):1290--1306, 2008.

\end{thebibliography}
}

\end{document}